\documentclass[runningheads]{llncs}

\usepackage{eccv}

\usepackage{eccvabbrv}

\usepackage{graphicx}
\usepackage{booktabs}

\usepackage[accsupp]{axessibility}  % Improves PDF readability for those with disabilities.

\usepackage[pagebackref,breaklinks,colorlinks,citecolor=eccvblue]{hyperref}
\usepackage{orcidlink}

\usepackage{multirow}  % for tables
\usepackage[dvipsnames]{xcolor} % for green text in tables.
\usepackage{subcaption}
\usepackage{xcolor}
\usepackage{microtype}
\usepackage{footnote}
\usepackage{tikz} % for shadow on table
\usetikzlibrary{shadows} % for shadow on table
\usepackage{amsmath} % writing the equations
\usepackage[ruled]{algorithm2e}
\usepackage{algpseudocode}
\usepackage{multirow} % for tables

\begin{document}

\title{Mixture of Efficient Diffusion Experts Through Automatic Interval and Sub-Network Selection}

\titlerunning{DiffPruning}

\author{Alireza Ganjdanesh\inst{1}\thanks{Part of this work was done during an internship at Adobe Research.} \and 
Yan Kang\inst{2} \and
Yuchen Liu\inst{2} \and
Richard Zhang\inst{2} \and
Zhe Lin\inst{2} \and
Heng Huang\inst{1}}

\authorrunning{A.~Ganjdanesh et al.}

\institute{Department of Computer Science, University of Maryland College Park \and
Adobe Research\\
\email{\{aliganj,heng\}@umd.edu,~\{yankang,yuliu,rizhang,zlin\}@adobe.com}} 

\maketitle

% \vspace{-0.26in}
\begin{abstract}
  Diffusion probabilistic models can generate high-quality samples.~Yet, their sampling process requires numerous denoising steps, making it slow and computationally intensive.~We propose to reduce the sampling cost by pruning a pretrained diffusion model into a mixture of efficient experts.~First, we study the similarities between pairs of denoising timesteps, observing a natural clustering, even across different datasets.~This suggests that rather than having a single model for all time steps, separate models can serve as ``experts'' for their respective time intervals.~As such, we separately fine-tune the pretrained model on each interval, with elastic dimensions in depth and width, to obtain experts specialized in their corresponding denoising interval.~To optimize the resource usage between experts, we introduce our Expert Routing Agent, which learns to select a set of proper network configurations.~By doing so, our method can allocate the computing budget between the experts in an end-to-end manner without requiring manual heuristics.~Finally, with a selected configuration, we fine-tune our pruned experts to obtain our mixture of efficient experts. We demonstrate the effectiveness of our method, \textit{DiffPruning}, across several datasets, LSUN-Church, LSUN-Beds, FFHQ, and ImageNet, on the Latent Diffusion Model architecture.
  \keywords{Efficient Deep Learning \and Model Pruning \and Diffusion Models}
  
\end{abstract}

\section{Introduction}
\label{sec:intro}

Diffusion Probabilistic Models (DPMs)~\cite{sohl2015nonequilibrium,ho2020ddpm,song2019-estimating-grads-of-data} have become the de facto models for generative modeling applications like image synthesis~\cite{ho2020ddpm,dhariwal2021DMsBeatGans}, image editing~\cite{zhang2023ControlNet,zhao2023Uni-ControlNet}, super-resolution~\cite{saharia2022SR3,gao2023ImplicitSuperResolution}, and video generation~\cite{ho2022imagen_video}.~They train a denoising model that learns to generate samples from an input noise in an iterative denoising scheme.~DPMs have achieved better mode coverage and training stability~\cite{dhariwal2021DMsBeatGans} than GANs~\cite{goodfellow2014GAN} and show higher sample quality than VAEs~\cite{KingmaVAE}.~Yet, the main drawback of DPMs is their slow and computationally intensive sampling process, making their cloud deployment costly and hindering usage on resource-constrained edge devices.

\begin{figure}[t]
\centering
\includegraphics[scale=0.32]{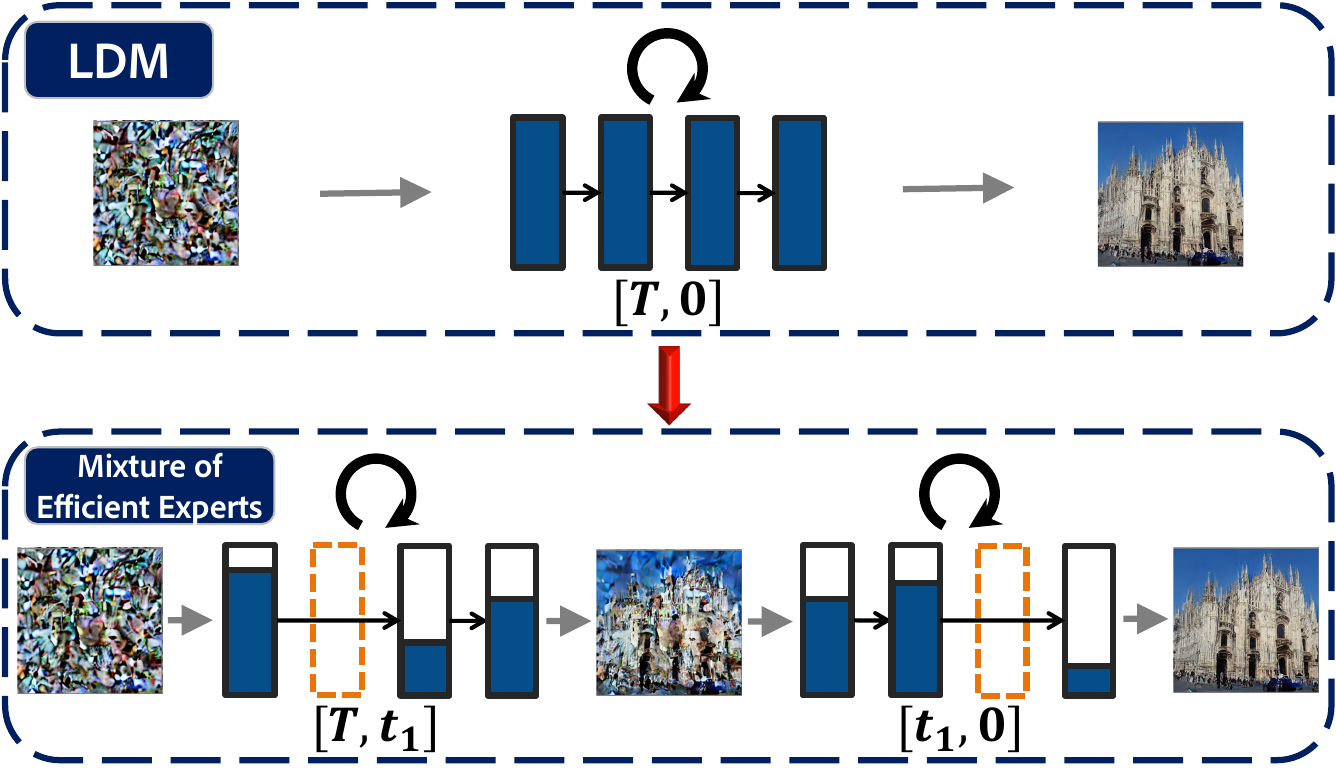}
\caption{\textbf{Overview of DiffPruning.} We prune a pre-trained LDM model~\cite{rombach2022LDM} (top) into a mixture of efficient experts (bottom). Each expert handles an interval, which allows their architectures to be separately specialized by removing layers or channels.}

\label{single_model_to_moe}
\end{figure}

% %-------------------------------------------------------------------------
% %-------------------------------------------------------------------------
Two important factors contribute to slow sampling in DPMs: the models use 1) a large number of denoising steps and 2) a large number of parameters in each denoising step.~Methods to speed up DPMs have primarily focused on reducing the sampling steps, using techniques like faster solvers~\cite{lu2022DPMSolver,bao2022AnalyticDPM,liu2022PseudoNumerical,zhang2023ExponentialIntegrator}, better noise schedules~\cite{song2021DDIM,nichol2021improvedDDPM}, and distillation~\cite{meng2023distillation,salimans2022ProgressiveDistillation,habibian2023clockwork}.~In an orthogonal direction, a group of methods address the second factor and develop more efficient architectures for DPMs.~Latent Diffusion Models~(LDMs)~\cite{rombach2022LDM} perform the diffusion process in a latent space with lower dimensions than pixel space, thereby significantly speeding up the sampling process while retaining a competitive performance.~Accordingly, LDMs have been deployed in modern generative models like DALL-E 3~\cite{betker2023dalle3} and Stable Diffusion~\cite{rombach2022LDM}.~Thus, compressing LDMs is of significant interest. As LDMs do not have redundancies of the pixel-space DPMs by design, pruning them is much more challenging than pruning pixel-space DPMs.

% %-------------------------------------------------------------------------
% %-------------------------------------------------------------------------

Recently, several works~\cite{lee2023MEME,liu2023OMS-DPM,zhang2023TailoredMultiDecoder} have explored architectural efficiency for LDMs.~They divide the denoising path of an LDM into several intervals and use a distinct model for each one.~These methods~\cite{lee2023MEME,liu2023OMS-DPM,zhang2023TailoredMultiDecoder} are mainly inspired by studies~\cite{balaji2022ediffiMOE,yang2023DPMsMadeSlim} showing different timesteps have distinct roles in the denoising process, and employing a single denoising model for all timesteps is sub-optimal~\cite{balaji2022ediffiMOE,go2023NegativeTransfer}.~Thus, the key design choices here are the clustering scheme of the denoising timesteps and the method for allocating the resource budget between the selected clusters.~MEME \cite{lee2023MEME} uses uniform clustering, and TMDA~\cite{zhang2023TailoredMultiDecoder} clusters the denoising timesteps by their loss values' similarities. Both MEME~\cite{lee2023MEME} and TDMA~\cite{zhang2023TailoredMultiDecoder} manually design a distinct U-Net model~\cite{ronneberger2015UNet} for each cluster, thereby heuristically allocating the resource budget between the denoising intervals.~However, by doing so, these methods need to re-design intervals' models for a new distinct budget, which is a complex, time-consuming, and labor-intensive task.~OMS-DPM~\cite{liu2023OMS-DPM} avoids manual designing intervals' models as it trains a model zoo with different sizes and searches for an optimal mixture of denoising models, given a desired computational budget.~Still, training a model zoo of various LDMs is extremely costly, even for medium-sized datasets, making OMS-DPM expensive to deploy in practice.

% %-------------------------------------------------------------------------
% %-------------------------------------------------------------------------

\begin{figure*}[t!]
  \centering
  \includegraphics[scale=0.23]{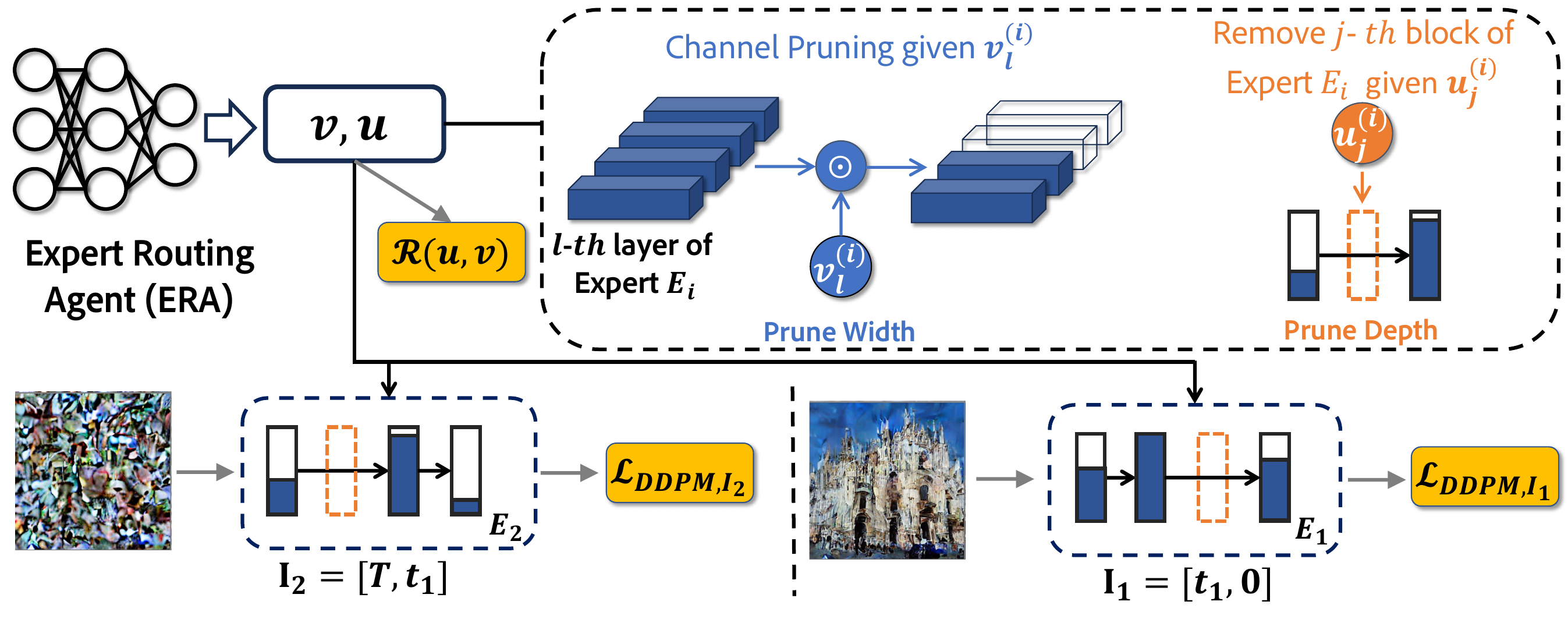}
  \caption{\textbf{Our Pruning Scheme:}~We train our Expert Routing Agent (ERA) to prune the experts into a mixture of efficient experts~(Sec.~\ref{ERA}).~The ERA predicts the architecture vectors $(v, u)$ to prune experts' width and depth.~Then, we calculate the denoising objectives of selected sub-networks of experts, $\mathcal{L}_{\text{DDPM},\mathcal{I}_i}$, as well as our Resource regularization term, $\mathcal{R}$, that encourages the ERA to provide a mixture of efficient experts with a desired compute budget (MACs).~We train ERA's parameters to minimize the objective functions. Thus, it learns to automatically allocate the compute budget (MACs) between experts in an end-to-end manner.}
  \label{pruning_scheme}
  % \vspace{-10pt}
\end{figure*}

In this paper, we propose a novel approach to make LDMs more efficient by pruning a large pretrained LDM into a mixture of efficient experts~(Fig.~\ref{single_model_to_moe}) in four steps.~First, we find an optimal division of time intervals by studying how aligned pairs of denoising steps are to each other in a pretrained LDM.~Interestingly, while different datasets all show natural clustering, the exact time intervals differ slightly between them. Thus, we adapt our clustering depending on the behavior of the dataset rather than using a static approach across datasets as in previous work~\cite{lee2023MEME}.
Second, we fine-tune the pretrained model with elastic depth and width on each interval so that the sub-networks of the resulting model have a strong performance on that interval.~We denote the elastically fine-tuned models as~\textit{experts} for the intervals.~Our elastic fine-tuning provides an `implicit' model zoo within each expert for its corresponding interval with fewer training iterations than training multiple models from scratch like OMS-DPM~\cite{liu2023OMS-DPM}.~Third, we develop an Expert Routing Agent (ERA) that learns to select proper network configurations for the experts simultaneously, guided by the sub-networks' denoising objectives and allocated compute resource (\textit{e.g.,} MACs).~As we train our ERA in an end-to-end manner, it can automatically allocate computing resources between the experts without the need for complex heuristics~\cite{lee2023MEME,zhang2023TailoredMultiDecoder}.~We summarize our contributions as follows:

\begin{itemize}
    \item We introduce a method for pruning LDMs into a mixture of efficient experts.
    \item We propose to cluster denoising timesteps of a pretrained LDM into several intervals based on their pairwise alignment scores,
    showing that the optimal clustering intervals are distinct for different datasets. We employ a specialized efficient model for each interval.
    \item We fine-tune the pretrained LDM on selected intervals with elastic dimensions so that resulting expert models will have strong sub-networks to choose from.~Thus, we can readily prune the experts for different computational budgets, and the pruned experts can properly recover their performance without long fine-tuning iterations.
    \item We develop a new pruning scheme in which our expert routing agent learns to select optimal layouts of the experts together in an end-to-end manner, thereby allocating the compute budget between experts automatically.
\end{itemize}

\section{Related Work}

\textbf{Mixture of Experts (MoE) diffusion models.} MoE methods cluster denoising timesteps of DPMs into intervals and train a separate~\textit{expert} model for each.~eDiff-I~\cite{balaji2022ediffiMOE} supports developing MoE for DPMs by showing that different denoising timesteps have separate roles.~Yet, how to cluster timesteps is non-trivial.~eDiff-I employs a tree-based-branching scheme, sequentially dividing the denoising path into two intervals and initializing a child model by its parent.~ERNIE-ViLG~\cite{feng2023ernieMOE} and MEME~\cite{lee2023MEME} uniformly cluster the denoising timesteps.~Yet, these heuristic schemes do not necessarily transfer to other tasks.~Alternatively,~we propose to cluster denoising timesteps in a data-driven way by measuring the alignment between their training objectives. We observe that the optimal cluster assignments are different for distinct datasets.~We note that although NT~\cite{go2023NegativeTransfer} has explored timesteps' alignment scores \textit{in the course of training}, our paper is the first one to leverage \textit{post-training scores} to cluster the timesteps for MoE DPMs.

% ------------------------------------------------------------------------------------------------------------------------------------------------

\noindent\textbf{Efficient DPMs.}~Ideas for improving DPMs' efficiency mostly reduce their denoising steps by faster samplers~\cite{lu2023dpmsolver++,zhang2023ExponentialIntegrator,xu2023RestartSampling}, distillation~\cite{salimans2022ProgressiveDistillation,meng2023distillation,habibian2023clockwork}, better noise schedules~\cite{zheng2023TruncatedDPMs,song2021DDIM,nichol2021improvedDDPM,zhang2023gddim,kingma2021variationalDMs,bao2022AnalyticDPM}, learning denoising steps~\cite{watson2022LearningFastSamplers,watson2022learning}, and caching~\cite{ma2023DeepCache}.~We explore an orthogonal direction, compressing DPMs' architectures.

A few ideas have recently addressed compressing DPMs' architectures having two main categories.~\textbf{Single-model} methods develop a single efficient model for all denoising timesteps.~SP~\cite{fang2023StructuralPruningforDMs} approximates weights' importance using the Taylor expansion and removes structures with low scores.~Yet, SP's performance has been mainly verified on pixel space DPMs, and its pruned models on datasets like LSUN-Church~\cite{yu15lsun} still have more than $6\times$ MACs than the full-size LDM~\cite{rombach2022LDM}.~MobileDiffusion~\cite{zhao2023MobileDiffusion} introduces heuristics to enhance DPMs' efficiency and develops two efficient architectures.~Nevertheless, it is highly non-trivial how to generalize the heuristics for different compute budgets.~Spectral Diffusion (SD)~\cite{yang2023DPMsMadeSlim} performs frequency domain distillation from a teacher model into a small LDM.~However, the main weakness of single-model methods is that they use the same model for all denoising steps, which is shown to be sub-optimal~\cite{balaji2022ediffiMOE,go2023NegativeTransfer}.~\textbf{Mixture of expert} methods employ a separate model for different stages of the denoising process.~OMS-DPM~\cite{liu2023OMS-DPM} trains a model zoo with various sizes and searches for a proper model schedule given a desired compute budget.~Yet, gathering a model zoo is very costly on large-scale datasets, making OMS-DPM impractical for them.~MEME~\cite{lee2023MEME} and TMDA~\cite{zhang2023TailoredMultiDecoder} cluster denoising timesteps and design a distinct expert for each.~However, they need to manually allocate the compute budget between experts and re-design the experts for a new budget, which makes them cumbersome in practice.~We prune an LDM into a mixture of efficient experts.~We cluster denoising steps into intervals using their alignment scores.~Then, we fine-tune the pre-trained model with elastic dimensions on each interval to obtain our experts.~Thus, our method gathers an \textit{implicit} model zoo \emph{within} each expert with much lower training iterations than OMS-DPM.~Finally, we prune all experts simultaneously using our expert routing agent to obtain our mixture of efficient experts.~By doing so, in contrast with MEME~\cite{lee2023MEME} and TMDA~\cite{zhang2023TailoredMultiDecoder}, our method automatically allocates the compute resource (\textit{e.g.,} MACs) between experts.~We refer to supplementary materials for a review of other related works.

% ------------------------------------------------------------------------------------------------------------------------------------------------

\section{Method}

We introduce a framework to prune an LDM~\cite{rombach2022LDM} model into a mixture of efficient experts in four steps.~First, we cluster denoising timesteps of the model into several intervals based on their objectives' alignment scores.~Second, we fine-tune the pre-trained model on the selected intervals with elastic dimensions to obtain our interval experts.~Third, we prune the experts together using our expert routing agent in an end-to-end manner (Fig.~\ref{pruning_scheme}).~Finally, we fine-tune the pruned experts to obtain our mixture of efficient experts.

% ------------------------------------------------------------------------------------------------------------------------------------------------
\begin{figure*}[t!]
  \centering
  \includegraphics[scale=0.36]{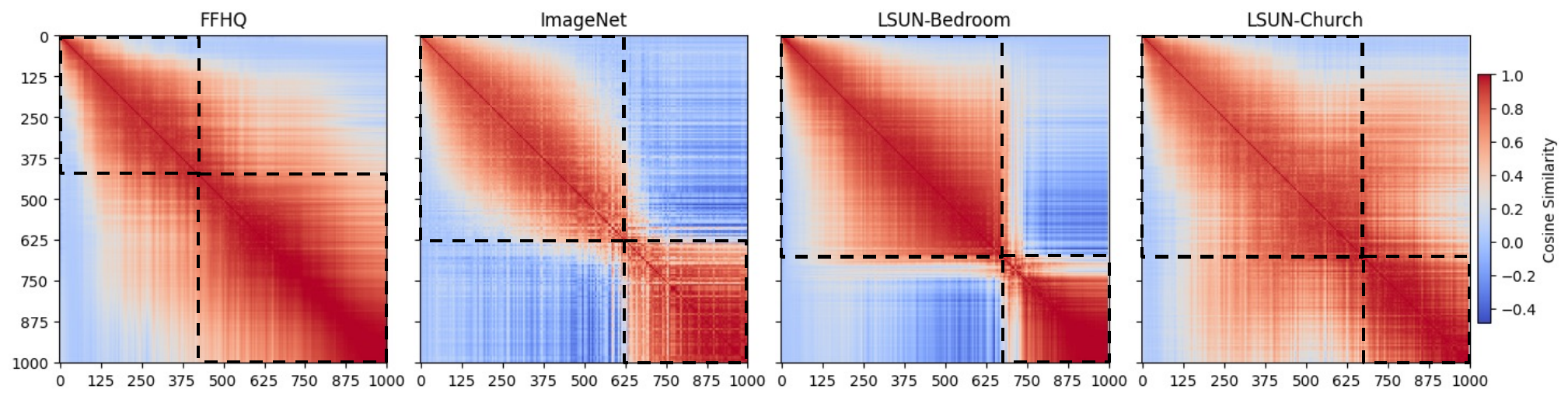}
  \caption{\textbf{Our Interval Selection Scheme:} We calculate gradients of denoising timesteps' objectives \textit{w.r.t} the pre-trained LDM's parameters and take the cosine similarity value of two timesteps' gradients as their alignment score.~The dashed lines show our selected cluster intervals for the experts.~One can observe the optimal cluster assignments are different for distinct datasets, and employing a deterministic clustering strategy~\cite{balaji2022ediffiMOE} like uniform clustering~\cite{feng2023ernieMOE} for all datasets is sub-optimal.}
  \label{ClusteringTimeSteps}
\end{figure*}

\subsection{Background}

Given a random variable $\mathbf{x_0} \sim \mathcal{P}$, the goal of DPMs~\cite{sohl2015nonequilibrium,ho2020ddpm} is to model the underlying distribution $\mathcal{P}$ using a training set $\mathcal{D}=\{x_0\}$ of samples.~To do so, first, DPMs define a forward process parameterized by $t$ in which they gradually perturb each sample $x_0$ with Gaussian noise with the variance schedule of $\beta_t$:

% \vspace{-10pt}
\begin{equation}
    q(x_t|x_{t-1}) = \mathcal{N}(x_t;\sqrt{1-\beta_t}x_{t-1}, \beta_tI)
\end{equation}

\noindent where $t\in[1, T].$ Thus, $q(x_t|x_0)$ has a Gaussian form:

% \vspace{-5pt}
\begin{equation}
    q(x_t|x_0) = \mathcal{N}(x_t;\sqrt{\Bar{\alpha_t}}x_{0}, (1-\Bar{\alpha_t})I)
\end{equation}

\noindent where $\alpha_t = 1 - \beta_t$ and $\Bar{\alpha}_t = \prod_{i=1}^{t}\alpha_i$.~The noise schedule $\beta_t$ is usually selected~\cite{ho2020ddpm} such that $q(x_T) \rightarrow \mathcal{N}(0, I)$.~Assuming $\beta_t$ is small, DPMs approximate the denoising distribution $q(x_{t-1}|x_t)$ by a parameterized Gaussian distribution $p_{\theta}(x_{t-1}|x_t) = \mathcal{N}(x_{t-1};\frac{1}{\sqrt{\alpha_t}}(x_t - \frac{\beta_t}{\sqrt{1 - \Bar{\alpha_t}}}\epsilon_{\theta}(x_t, t)), \sigma^2_tI)$, and~$\sigma^2_t$ is often set to $\beta_t$.~DPMs implement $\epsilon_{\theta}(.)$ with a neural network called the denoising model and train it with the variational evidence lower bound (ELBO) objective~\cite{ho2020ddpm}:

\begin{equation}\label{denoising-obj}
    \begin{split}
        \mathcal{L}_{\text{DDPM}} &= \mathbb{E}_{\substack{t\sim[1, T]}}\mathcal{L}_t \\ &= \mathbb{E}_{{t\sim[1, T], \epsilon\sim\mathcal{N}(0, I), x_t\sim q(x_t|x_0)}}||\epsilon_{\theta}(x_t, t) - \epsilon||^2     
    \end{split}
\end{equation}

\noindent DPMs generate a new sample by sampling an initial noise from $x_T\sim p(x_T)=\mathcal{N}(0, I)$ and iteratively denoising it using the denoising model by sampling from $p_{\theta}(x_{t-1}|x_t)$.~Thus, the sampling process requires $T$ sequential forward passes to the large denoising model, making it a slow and costly process.

\subsection{Notations}\label{notations}
\begin{figure}[t!]
\centering
\includegraphics[scale=0.4]{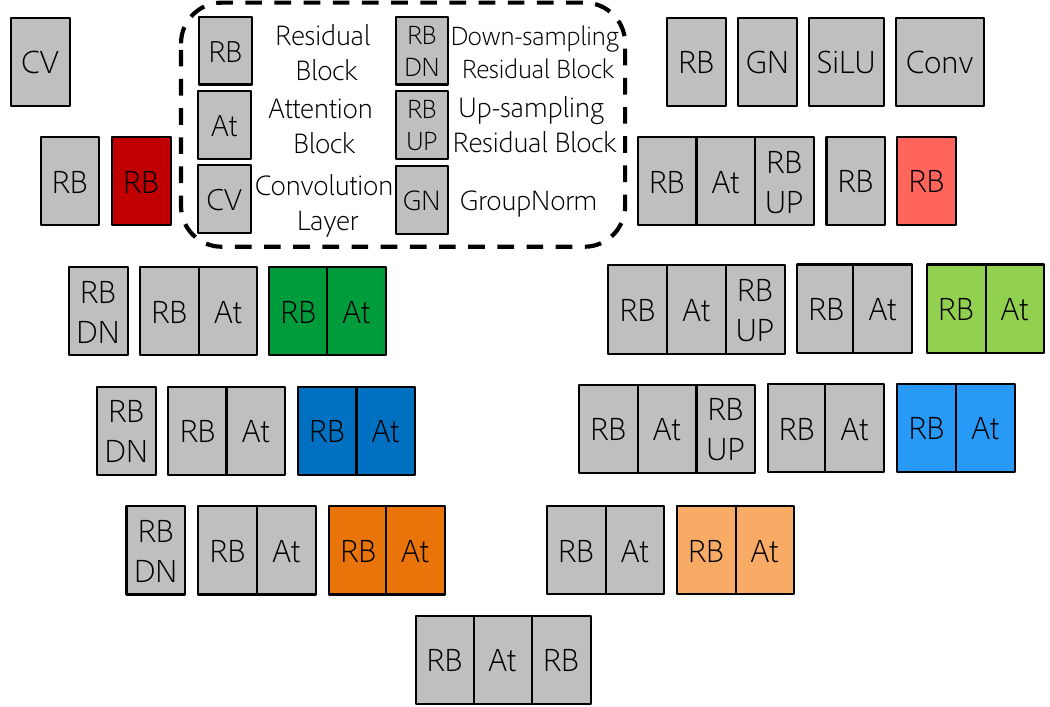}
\caption{U-Net architecture of the LDM~\cite{rombach2022LDM}.~We randomly drop/preserve each colored layer in our elastic depth fine-tuning.}
\label{depth_pruning_blocks}
\end{figure}

Fig.~\ref{depth_pruning_blocks} shows the U-Net~\cite{ronneberger2015UNet} architecture used in LDM~\cite{rombach2022LDM} models.~The encoder and decoder branches have several \textit{stages} (each row in Fig.~\ref{depth_pruning_blocks}).~Each stage has one or several~\textit{layers}.~We represent the layers' functions and feature~maps with $f_{l}(.)$ and $\mathcal{F}_{l}$, respectively, where $l\in[1, L]$, and $L$ is the total number of layers.~Each layer consists of one or several~\textit{blocks}.~For instance, the green-colored \textit{layers} in Fig.~\ref{depth_pruning_blocks} are in the third \textit{stage} of its encoder and decoder and consist of a Residual~\textit{block}~\cite{he2016ResNet} and an Attention~\textit{block}~\cite{vaswani2017Attention}.

\subsection{Clustering Denoising Timesteps into Intervals}\label{experts-intervals}
We propose to cluster denoising timesteps $\mathcal{T}=[1, T]$ of an LDM into $N$ intervals $\{\mathcal{I}_1, \mathcal{I}_2, \cdots, \mathcal{I}_N\}$ such that $\mathcal{T}=\mathcal{I}_1\cup\mathcal{I}_2\cup\cdots\cup\mathcal{I}_N$.~The intuition is that different intervals have separate roles~\cite{balaji2022ediffiMOE}.~For example, it has been empirically shown~\cite{choi2022PerceptionPrioritized} that an LDM first generates the layout of an image in high-noise timesteps and then fills in the details in low-noise ones.~Thus, using the same denoising model for all timesteps is sub-optimal.~We employ alignment scores of training objectives $\mathcal{L}_t$~(Eq.~\ref{denoising-obj}) for denoising timesteps of a pre-trained LDM to cluster them.~We estimate the gradient of each $\mathcal{L}_t$ \textit{w.r.t} the denoising model's parameters~($\theta$) using a random batch of samples in the training data and take the cosine similarity between the gradients of $\mathcal{L}_t$ and $\mathcal{L}_s$ as the alignment score of timesteps $t$ and $s$.

We visualize pairwise alignment scores of denoising time-steps for pre-trained LDMs~\cite{rombach2022LDM} of different datasets in Fig.~\ref{ClusteringTimeSteps}.~We select two distinct clusters ($N$ = $2$) for all datasets (shown by dashed lines in Fig.~\ref{ClusteringTimeSteps}) in our experiments.~We choose the cut-off point between the clusters to be the one that maximizes the weighted mean of the average scores of the clusters, and we refer to the supplementary for the formulation.~We do not use more than two experts in our experiments for computational efficiency.~However, our formulation can find cut-off points for more than two clusters, as we elaborate in the supplementary.~One can observe that intra-cluster alignment scores are high, and inter-cluster scores are small and even negative for LSUN-Bedroom~\cite{yu15lsun}, ImageNet~\cite{RussakovskyImageNet}, and FFHQ~\cite{karras2019StyleGAN_FFHQ}. Accordingly, further training the denoising model on one of the clusters degrades its performance on the other.~This observation supports our decision to employ a specialized model for each interval, that using a single model for all intervals is sub-optimal.~Further, the optimal cluster assignment is different for distinct datasets.~Thus, our clustering method is more robust than deterministic ones~\cite{feng2023ernieMOE,balaji2022ediffiMOE}.~In summary, we select clusters $(\mathcal{I}_1, \mathcal{I}_2)$ to be $([0, 700], [701, 1000])$ for LSUN-Church as well as LSUN-Bedroom,~$([0, 400], [401, 1000])$ for FFHQ, and $([0, 625], [626, 1000])$ for ImageNet.

\subsection{Fine-tuning with Elastic Dimensions}
\label{elastic_finetuning}

We fine-tune the pre-trained LDM with elastic dimensions (depth and width) on each denoising interval $\mathcal{I}_i~(i\in[1,N])$ after clustering the denoising timesteps.~We call the resulting models \textit{experts} and denote them with $E_i$, corresponding to $\mathcal{I}_i$.~Our main inspiration is that by doing so, each sub-network of an expert $E_i$ has a decent performance on the denoising interval $\mathcal{I}_i$, which brings in several key benefits:~First, the loss value of each sub-network of $E_i$ will be a proper proxy for its actual performance after fine-tuning on $\mathcal{I}_i$.~Second, the pruned experts will be able to recover their performance promptly during fine-tuning without requiring long fine-tuning iterations.~Finally, our elastic fine-tuning provides a model zoo \textit{within} the expert $E_i$ for the denoising interval $\mathcal{I}_i$ without extreme computational and memory expensive training of several architectures from scratch like in OMS-DPM~\cite{liu2023OMS-DPM}.~We first fine-tune the pre-trained model with elastic depth on each interval.~Then, we fine-tune the resulting model with elastic width.~We do not perform elastic depth and width training together to prevent instabilities.

\noindent\textbf{Fine-tuning with elastic depth.}~We randomly drop the last layer in each stage of the U-Net's encoder and decoder (colored blocks in Fig.~\ref{depth_pruning_blocks}) for our elastic depth training.~Formally, for each training batch, we randomly select to map the last depth layers $f_{j}^{(i)}$ in stages of the expert $E_i$ independently with a probability $p$ to the identity function:

\begin{equation}\label{elastic_depth}
    \small\hat{f}_{j}^{(i)}=f_{j}^{(i)}\textbf{1}_{\{s_j^{(i)}=0\}} + I\textbf{1}_{\{s_j^{(i)}=1\}},~~~ s_j^{(i)}\sim \text{Bernoulli}(p)
\end{equation}

\noindent We train selected sub-network's parameters with the interval denoising objective:

\begin{equation}\label{interval_denoising_obj}
    \mathcal{L}_{\text{DDPM},\mathcal{I}_i} = \mathbb{E}_{t\sim\mathcal{I}_i}\mathcal{L}_t
\end{equation}

\noindent where $\mathcal{L}_t$ has the same formulation as Eq.~\ref{denoising-obj}.

%%%%%%%%%%%%%%%%%%%%%%%%%%%%%%%%%%%%%%%%%%%%%%%%%%%%%%%%%%%%%%%%%%%%%%%%%%%%%%%%%%%%%%%%%%%%%%%%%%%%%%%%%%%%%%%%%%%%%%%%%%%%%%%%%%%%%%%%%%%%%%%%%%%%%%%%%%%%%%%%%%%%%%%%%%%%%%%%%%%%%%%%%%%%%%%%%%%%%%%%%%%%%%%%%%%%%%%%%%%%%%%%%%%%

\noindent\textbf{Fine-tuning with elastic width.}~After fine-tuning the pre-trained model on each interval $\mathcal{I}_i$ with elastic depth, we fine-tune the resulting experts with elastic width.~For each ResBlock in the U-Net, we sort the channels of its convolution layers based on an estimate of their importance (determined by their $L_1$ norm~\cite{li2017L1NormPruning,Cai2020OFA}).~Then, for each training batch, we randomly remove some ratio of the least important channels of each convolution layer.~Similarly, for the attention layers~\cite{vaswani2017Attention}, we sort the attention heads based on the $L_1$ norm of their projection weights, and we randomly drop some of the least important heads during our elastic width fine-tuning.~Finally, we update the selected sub-network's weights using $\mathcal{L}_{\text{DDPM},\mathcal{I}_i}$ (Eq.~\ref{interval_denoising_obj}).~We refer to supplementary for more details.

\subsection{Expert Routing Agent}\label{ERA}
We develop our Expert Routing Agent (ERA) to prune the elastically fine-tuned experts $E_i~(i\in[1, N])$ into a mixture of efficient experts.~We denote our ERA as a function $h_\text{ERA}(.;\beta)$ parameterized by $\beta$, predicting architecture vectors $(u, v)$:

% \vspace{-5pt}
\begin{equation}
    u, v = h_\text{ERA}(z; \beta)
\end{equation}

\noindent $z$ is a constant, randomly initialized input.~Vectors $u = [u^{(i)}]_{i=1}^{N}$ determine pruning depth layers $f_{j}^{(i)}$ (Eq.~\ref{elastic_depth}).~Similarly, vectors $v = [v^{(i)}]_{i=1}^{N}$ determine widths of blocks of $N$ experts. Together, $(u,v)$ select sub-networks $e_i$ from experts $E_i$.

Given a total constraint on the computation budget (\textit{e.g.}, MACs, latency, etc.), denoted as $T_d$, we optimize the ERA model's parameters $\beta$ to predict architecture vectors $(u, v)$ for an efficient and high-performing set of experts' architectures. Next, we describe how we parameterize and apply $(u,v)$. We show the formulation to determine the compute budget of selected architectures and the final optimization procedure for the ERA in Eq. \ref{eqn:era_optim}.

\subsubsection{Pruning Width:} Although one can prune widths of blocks in an expert $E_i$ using a binary vector $v^{(i)}$, keeping the \textit{j}$^\text{th}$ channel when $v_j^{(i)}$ is $1$ and vice versa, such an operation is not differentiable, making the optimization of parameters $\beta$ of the ERA challenging.~Thus, we introduce soft vectors $\mathbf{v}^{(i)}$, relax them to have continuous values, and use them for width pruning. We calculate them as follows:

\begin{equation}\label{width-gumbell}
    \mathbf{v}^{(i)} = \text{sigmoid}(\frac{v^{(i)} + n}{\tau})
\end{equation}

\noindent $n\sim\text{Gumbel}(0, 1)$ is a noise from the Gumbel distribution~\cite{gumbel1954GumbellDistribution}. Parameter $\tau$ is the temperature that when set appropriately, brings elements of $\mathbf{v}^{(i)}$ close to $0$ or $1$.~The calculation from $v^{(i)}$ to $\mathbf{v}^{(i)}$ is called the Gumbel-Sigmoid trick~\cite{maddison2017ConcreteDistribution,jang2017GumbelSoftmax}.~It is a differentiable estimation of sampling from a Bernoulli distribution with the Bernoulli parameter of $\texttt{sigmoid(}v^{(i)}\texttt{)}$.~We apply vectors $\mathbf{v}^{(i)}=[\mathbf{v}_{l}^{(i)}]_{l=1}^{L}$ to prune the width of blocks in all of the layers $f_{l}^{(i)}$ for the expert $E_i$:

\begin{equation}\label{width_pruning_eq}
    \hat{f}_{l}^{(i)} = f_{l}^{(i)}(.; \mathbf{v}_{l}^{(i)})
\end{equation}

\noindent Here, we apply (multiply) the width vector $\mathbf{v}_{l}^{(i)}$ to feature maps of the first convolution layer in the ResBlocks and inputs of the attention operation in the attention blocks in the layer $f_{l}^{(i)}$.~The granularity of our width pruning is similar to our elastic width fine-tuning,~\textit{i.e.,} we prune channels of convolution layers of ResBlocks and heads of the attention layers.

%%%%%%%%%%%%%%%%%%%%%%%%%%%%%%%%%%%%%%%%%%%%%%%%%%%%%%%%%%%%%%%%%%%%%%%%%%%%%%%%%%%%%%%%%%%%%%%%%%%%%%%%%%%%%%%%%%%%%%%%%%%%%%%%%%%%%%%%%%%%%%%%%%%%%%%%

\subsubsection{Pruning depth.} Similarly, we employ relaxed continuous vectors $\mathbf{u}^{(i)}=[\mathbf{u}_{j}^{(i)}]$ for pruning the depth layers $f_{j}^{(i)}$ (Eq.~\ref{elastic_depth}) of the expert $E_i$:

\begin{equation}\label{depth-gumbell}
    \mathbf{u}^{(i)} = \text{sigmoid}(\frac{u^{(i)} + n}{\tau})
\end{equation}

As there are skip connections in the U-Net, we apply the depth architecture vectors for the encoder and decoder branches differently.

\noindent\textbf{Encoder depth pruning.} We use the following formulation to apply the vector $\mathbf{u}^{(i)}$ for pruning depth layers in the encoder of the expert $E_i$:

% \vspace{-5pt}
\begin{equation}\label{depth_pruning_enc}
    \mathcal{\widehat{F}}_{j}^{(i)} = \mathbf{u}_{j}^{(i)}f_{j}^{(i)}(\mathcal{F}_{j-1}^{(i)}) + (1 - \mathbf{u}_{j}^{(i)})\mathcal{F}_{j-1}^{(i)}
\end{equation}

\noindent In other words, we interpolate between the feature map of the previous layer, $\mathcal{F}_{j-1}^{(i)}$, and the result of applying the current layer to it, $f_{j}^{(i)}(\mathcal{F}_{j-1}^{(i)})$.~The $\mathbf{u}_{j}^{(i)}$ values close to $1$ simulate preserving the layer, and $0$ simulate removing the layer.

\noindent\textbf{Decoder depth pruning.} The input for the layer $f_{j}^{(i)}$ in the decoder of the expert $E_i$ is the concatenation of feature maps $\mathcal{F}_{j-1}^{(i)}$ of its previous layer and the skip connection feature maps $\mathcal{F}_{j, skip}^{(i)}$.~Thus, we apply the vector $\mathbf{u}_{j}^{(i)}$ to it as:

\begin{equation}\label{depth_pruning_dec}
    \mathcal{\widehat{F}}_{j}^{(i)} = \mathbf{u}_{j}^{(i)}f_{j}^{(i)}(\mathcal{F}_{j-1}^{(i)}\mathbin\Vert\mathcal{F}_{j, skip}^{(i)}) + (1 - \mathbf{u}_{j}^{(i)})\mathcal{F}_{j-1}^{(i)}
\end{equation}

\noindent where $\mathbin\Vert$ denotes concatenation.~Similar to Eq.~\ref{depth_pruning_enc}, we interpolate between applying or removing the layer in Eq.~\ref{depth_pruning_dec}.

%%%%%%%%%%%%%%%%%%%%%%%%%%%%%%%%%%%%%%%%%%%%%%%%%%%%%%%%%%%%%%%%%%%%%%%%%%%%%%%%%%%%%%%%%%%%%%%%%%%%%%%%%%%%%%%%%%%%%%%%%%%%%%%%%%%%%%%%%%%%%%%%%%%%%%%%
% \vspace{-10pt}
\subsection{Pruning the Mixture of Experts}
We train our Expert Routing Agent to select competent sub-networks of elastically trained experts given a desired total compute budget.~We measure the compute budget of our models with MACs, following~\cite{lee2023MEME,fang2023StructuralPruningforDMs}.~Given an architecture width vector $\textbf{v}_{l}^{(i)}$, the MACs of the layer $f_{l}^{(i)}(.)$ after applying $\textbf{v}_{l}^{(i)}$ will be:

\begin{equation}
    \widehat{T}_{l}^{(i)} = \textbf{1}^T \times \lfloor\textbf{v}_{l}^{(i)}\rceil \times T_{l}^{(i)}\label{width_MACs}
\end{equation}

\noindent where \textbf{1} denotes a vector of all ones. $\lfloor\cdot\rceil$ is the function that rounds to the nearest integer, and $T_{l}^{(i)}$ is the MACs of the layer $f_{l}^{(i)}(.)$.~Similarly, the MACs for the layers $f_{j}^{(i)}(.)$ that we use for depth pruning (Eq.~\ref{elastic_depth}) after applying $\textbf{u}_{j}^{(i)}$ will be:

\begin{equation}
    \widehat{T}_{j}^{(i)} = \lfloor\textbf{u}_{j}^{(i)}\rceil \times \textbf{1}^T \times \lfloor\textbf{v}_{j}^{(i)}\rceil \times T_{j}^{(i)} \label{depth_MACs}
\end{equation}

\noindent After applying architecture vectors of each expert $E_i$, we calculate the total MACs of our mixture of experts as:

\begin{equation}\label{total-macs}
    \widehat{T}(u, v) = \sum_{i=1}^{N} \frac{\left|\mathcal{I}_i\right|}{\sum_{k=1}^{N} \left|\mathcal{I}_k\right|} \widehat{T}^{(i)}(u^{(i)}, v^{(i)}) 
\end{equation}

\noindent where $\widehat{T}^{(i)}(u^{(i)}, v^{(i)})$ is the MACs of the expert $E_i$ after applying its architecture width and depth vectors.~In Eq.~\ref{total-macs}, we assume that the denoising schedule is \textit{linear} such that the number of denoising steps that the expert $E_i$ will contribute to the denoising process is proportional to the size of its interval $\left|\mathcal{I}_i\right|.$ One can alter Eq.~\ref{total-macs} for other denoising schedules like quadratic~\cite{song2021DDIM}, but we focus on the linear schedule as it has been widely adopted in the literature~\cite{song2021DDIM,rombach2022LDM,lee2023MEME,fang2023StructuralPruningforDMs}.

Given a desired MACs budget $T_d$, we train our ERA with the following objective to encourage it to select sub-networks of experts such that each of them has a high performance and their mixture has a total MACs close to $T_d$:

% \vspace{-5pt}
\begin{equation}
\label{eqn:era_optim}
        \small\min_{\beta}\mathcal{J}(T_d) = [\frac{1}{N}\sum_{i=1}^{N}\mathcal{L}_{\text{DDPM}, \mathcal{I}_i}(E_i(u^{(i)}, v^{(i)}))] + \mathcal{R}(\widehat{T}(u, v), T_d)
\end{equation}

\noindent $\mathcal{L}_{\text{DDPM}, \mathcal{I}_i}(E_i(u^{(i)}, v^{(i)}))$ is the interval denoising objective (Eq.~\ref{interval_denoising_obj}) of the sub-network of $E_i$ chosen by $(u^{(i)}, v^{(i)})$.~$\widehat{T}(u,v)$ is the total MACs of the mixture of experts~(Eq.~\ref{total-macs}) determined by the architecture vectors $(u,v)$ that are functions of the ERA's parameters $\beta$.~$\mathcal{R}(\cdot)$ is the MACs regularization term that we implement it as $\mathcal{R}(x,y) = \log(\max(x,y)/\min(x, y))$.~Now, as the round function $\lfloor\cdot\rceil$ used in Eqs.~(\ref{width_MACs},~\ref{depth_MACs}) is not differentiable, we use the  Straight Through Estimator (STE)~\cite{bengio2013STE} to calculate the gradients of $\mathcal{R}$ \textit{w.r.t} the parameters $\beta$ of our ERA.~We implement our ERA model with a GRU~\cite{ChoGRU} layer followed by dense layers.~We found in our experiments that a lightweight ($\sim0.5M$ parameters) ERA model suffices to obtain a performant mixture of efficient experts.~We show our pruning scheme in Fig.~\ref{pruning_scheme} and refer to supplementary for more details of our pruning algorithm as well as the ERA's architecture.

\subsubsection{Fine-tuning pruned models.}After our pruning stage, we use architecture vectors predicted by the ERA to prune experts.~Then, we fine-tune the experts with the same settings as the original LDM model~\cite{rombach2022LDM}.

\section{Experiments}

We experiment on the LSUN-Church~\cite{yu15lsun},~LSUN-Bedroom~\cite{yu15lsun},~FFHQ~\cite{karras2019StyleGAN_FFHQ},~and class-conditional ImageNet~\cite{RussakovskyImageNet} to verify our method's effectiveness.~We apply our method to the LDM~\cite{rombach2022LDM} that implements their denoising model with a U-Net~\cite{ronneberger2015UNet} architecture.~For all datasets, we prune the LDM model with three MACs budgets of 70$\%$,~50$\%$,~and 30$\%$.~In all experiments, we mainly follow the same hyper-parameter settings as LDM~\cite{rombach2022LDM}, and we refer to supplementary for more details of our experimental setup.~We denote our method as DiffPruning in the rest of the experiments section. We mainly compare our method with a few recent baselines on architectural efficiency of pixel-space~\cite{fang2023StructuralPruningforDMs} and latent space~\cite{liu2023OMS-DPM,lee2023MEME} DPMs.~We do not benchmark with Spectral Diffusion (SD)~\cite{yang2023DPMsMadeSlim} because it uses a package\footnote{\href{https://github.com/sovrasov/flops-counter.pytorch}{https://github.com/sovrasov/flops-counter.pytorch}} that does not count MACs of the attention operation $\texttt{QKVAttention}$ implemented in the LDM repository\footnote{\href{https://github.com/CompVis/latent-diffusion/blob/a506df5756472e2ebaf9078affdde2c4f1502cd4/ldm/modules/diffusionmodules/openaimodel.py\#L379}{https://github.com/CompVis/latent-diffusion}}.~Hence, the MACs of models reported by SD~\cite{yang2023DPMsMadeSlim} are not accurate.~For instance, SD reports the LDM~\cite{rombach2022LDM} for LSUN-Church has 18.7G MACs, but it actually has 20.96G~(Tab.~\ref{lsun_church_results}).

\subsection{Comparison Results}
We summarize comparison results in Tab.~\ref{all_results} and refer to supplementary for FID~\emph{vs.}~MACs as well as FID~\emph{vs.}~Throughput curves of our method and baselines.

%%%%%%%%%%%%%%%%%%%%%%%%%%%%%%%%%%%%%%%%%%%%%%%%%%%%%%%%%%%%%%%%%%%%%%%%%%%%%%%%%%%%%%%%%%%%%%%%%%%%%%%%%%%%%%%%%%%%%%%%%%%%%%%%%%%%%%%%%%%%%%%%%%%%%%%%%%%%%%%%%%%%%%%%%%%%%%%%%%%%%%%%%%%%%%%%%%%%%%%%%%%%%%%%%%%%%%%%%%%%%%
% LSUN-Bedroom

\noindent\textbf{LSUN-Bedroom.}~Tab.~\ref{lsun_bed_results} presents the results on LSUN-Bedroom.~First, we can observe that the LDM~\cite{rombach2022LDM} can achieve better sample quality (lower FID) while having significantly lower MACs (higher sampling speed) than the pixel-space DPM,~DDPM~\cite{ho2020ddpm}.~Although SP~\cite{fang2023StructuralPruningforDMs} prunes more than 44$\%$ MACs of the DDPM~\cite{ho2020ddpm}, its pruned model still has $36\%$ more MACs than LDM while drastically degrading the sample quality to 18.6 FID.~These results demonstrate that pixel-space DPMs have much more redundancies than LDMs, and pruning LDMs is significantly more challenging.~Second, our pruned models can achieve higher throughput speed-up ratio than their pruning ratio while having competitive FID scores.~DiffPruning $70\%$/$50\%$/$30\%$ models reduce MACs by $30\%$/$50\%$/$70\%$, but can secure $43\%$/$87\%/135\%$ sampling speed up compared to LDM~\cite{rombach2022LDM}.~Notably, DiffPruning $70\%$~($50\%$) has 5.90~(6.73) FID score which is fairly close to the original LDM (4.39) while substantially better than the pruned model by SP~\cite{fang2023StructuralPruningforDMs} (18.6).~Finally, although not directly comparable, our pruned models require less training iterations than SP, and we refer to supplementary for details.

\begin{table*}[t!] % [t] means "top of the page"
    \centering
    \caption{Comparison results of DiffPruning \emph{vs.} baselines.~Throughput values are calculated using an NVIDIA A100 GPU.~$\dagger$: the values are average of our two efficient experts.~$*$: calculated by sampling from provided checkpoints.~$\ddagger$: speed-ups relative to the LDM model.~The shadowed values are inaccurate, and we refer to supplementary for a detailed discussion.}
    \begin{subtable}{0.494\linewidth}
    \resizebox{\linewidth}{!}{
        \begin{tabular}{ccccc}
        \hline
        \multicolumn{5}{c}{LSUN-Bedroom ($256\times256$)}                                                                                                                                                    \\ \hline
        \multicolumn{1}{c|}{}             & \multicolumn{2}{c|}{Complexity}                             & \multicolumn{2}{c}{Performance}                                                               \\ \hline
        \multicolumn{1}{c|}{Model}        & \multicolumn{1}{c|}{Params}  & \multicolumn{1}{c|}{MACs}    & \multicolumn{1}{c|}{\begin{tabular}[c]{@{}c@{}}Throughput $(\uparrow)$\\ (Sample/Sec)\end{tabular}} & FID $(\downarrow)$  \\ \hline
        \multicolumn{1}{c|}{DDPM~\cite{ho2020ddpm}}         & \multicolumn{1}{c|}{113.7M}  & \multicolumn{1}{c|}{248.7G}  & \multicolumn{1}{c|}{0.74}                                                              & 6.62 \\
        \multicolumn{1}{c|}{SP~\cite{fang2023StructuralPruningforDMs}}           & \multicolumn{1}{c|}{63.2M}   & \multicolumn{1}{c|}{138.8G}  & \multicolumn{1}{c|}{-}                                                                 & 18.6 \\
        \multicolumn{1}{c|}{LDM~\cite{rombach2022LDM}}          & \multicolumn{1}{c|}{274.06M} & \multicolumn{1}{c|}{101.32G} & \multicolumn{1}{c|}{2.01}                                                              & 4.39$^*$ \\ \hline
        \multicolumn{1}{c|}{DiffPruning (70\%)} & \multicolumn{1}{c|}{162.06M$^\dagger$} & \multicolumn{1}{c|}{70.84G}  & \multicolumn{1}{c|}{\textbf{3.11}~\textbf{\color{ForestGreen}($\times$1.55)}$^\ddagger$}                                                              & 5.90 \\
        \multicolumn{1}{c|}{DiffPruning (50\%)} & \multicolumn{1}{c|}{100.87M$^\dagger$} & \multicolumn{1}{c|}{50.69G}  & \multicolumn{1}{c|}{\textbf{3.75}~\textbf{\color{ForestGreen}($\times$1.87)}$^\ddagger$}                                                              & 6.73 \\
        \multicolumn{1}{c|}{DiffPruning (30\%)} & \multicolumn{1}{c|}{48.43M$^\dagger$}  & \multicolumn{1}{c|}{31.11G}  & \multicolumn{1}{c|}{\textbf{4.73}~\textbf{\color{ForestGreen}($\times$2.35)}$^\ddagger$}                                                              & 9.22 \\ \hline
        \end{tabular}
    }
    \caption{}
    \label{lsun_bed_results}
    \end{subtable}
    \hspace{0.0em} % Space between tables
    \begin{subtable}{0.49\linewidth}
    \resizebox{\linewidth}{!}{
        \begin{tabular}{ccccc}
        \hline
        \multicolumn{5}{c}{FFHQ ($256\times256$)}                                                                                                                                                             \\ \hline
        \multicolumn{1}{c|}{}             & \multicolumn{2}{c|}{Complexity}                             & \multicolumn{2}{c}{Performance}                                                                \\ \hline
        \multicolumn{1}{c|}{Model}        & \multicolumn{1}{c|}{Params}  & \multicolumn{1}{c|}{MACs}    & \multicolumn{1}{c|}{\begin{tabular}[c]{@{}c@{}}Throughput $(\uparrow)$\\ (Sample/Sec)\end{tabular}} & FID $(\downarrow)$   \\ \hline
        \multicolumn{1}{c|}{LDM~\cite{rombach2022LDM}}          & \multicolumn{1}{c|}{274.06M} & \multicolumn{1}{c|}{101.32G} & \multicolumn{1}{c|}{1.01}                                                              & 9.53$^*$  \\
        \multicolumn{1}{c|}{DiffPruning (70\%)} & \multicolumn{1}{c|}{194.79M$^\dagger$} & \multicolumn{1}{c|}{71.05G}  & \multicolumn{1}{c|}{\textbf{1.35} \textbf{\color{ForestGreen}($\times$1.33)}$^\ddagger$}                                                              & 9.80  \\
        \multicolumn{1}{c|}{DiffPruning (50\%)} & \multicolumn{1}{c|}{134.67M$^\dagger$} & \multicolumn{1}{c|}{51.87G}  & \multicolumn{1}{c|}{\textbf{1.83} \textbf{\color{ForestGreen}($\times$1.81)}$^\ddagger$}                                                              & 9.90  \\
        \multicolumn{1}{c|}{DiffPruning (30\%)} & \multicolumn{1}{c|}{63.07M$^\dagger$}  & \multicolumn{1}{c|}{30.68G}  & \multicolumn{1}{c|}{\textbf{2.90} \textbf{\color{ForestGreen}($\times$2.87)}$^\ddagger$}                                                              & 10.66 \\ \hline
        \end{tabular}
    }
    \caption{}
    \label{ffhq_results}
    \end{subtable}
    \\[0ex] % Add some vertical space between rows of tables
    \begin{subtable}{0.503\linewidth}
    \resizebox{\linewidth}{!}{
        \begin{tabular}{cccccc}
        \hline
        \multicolumn{6}{c}{LSUN-Church ($256\times256$)}                                                                                                                                                                                                       \\ \hline
        \multicolumn{1}{c|}{}             & \multicolumn{3}{c|}{Complexity}                                                                              & \multicolumn{2}{c}{Performance}                                                                \\ \hline
        \multicolumn{1}{c|}{Model}        & \multicolumn{1}{c|}{Sampler}                   & \multicolumn{1}{c|}{Params}   & \multicolumn{1}{c|}{MACs}   & \multicolumn{1}{c|}{\begin{tabular}[c]{@{}c@{}}Throughput $(\uparrow)$\\ (Sample/Sec) \end{tabular}} & FID $(\downarrow)$   \\ \hline
        \multicolumn{1}{c|}{LDM~\cite{rombach2022LDM}}          & \multicolumn{1}{c|}{DDIM-100}                  & \multicolumn{1}{c|}{294.7M}   & \multicolumn{1}{c|}{20.96G} & \multicolumn{1}{c|}{5.19}                                                              & 5.21$^*$  \\
        \multicolumn{1}{c|}{LDM~\cite{rombach2022LDM}}          & \multicolumn{1}{c|}{DDIM-200}                  & \multicolumn{1}{c|}{294.7M}   & \multicolumn{1}{c|}{20.96G} & \multicolumn{1}{c|}{2.60}                                                              & 5.11$^*$  \\ \hline
        \multicolumn{1}{c|}{DDPM~\cite{song2021DDIM}}         & \multicolumn{1}{c|}{\multirow{3}{*}{DDIM-100}} & \multicolumn{1}{c|}{113.7M}   & \multicolumn{1}{c|}{248.7G} & \multicolumn{1}{c|}{0.74}                                                              & 10.58 \\
        \multicolumn{1}{c|}{SP~\cite{fang2023StructuralPruningforDMs}}           & \multicolumn{1}{c|}{}                          & \multicolumn{1}{c|}{63.2M}    & \multicolumn{1}{c|}{138.8G} & \multicolumn{1}{c|}{-}                                                                 & 13.9  \\
        \multicolumn{1}{c|}{DiffPruning (70\%)} & \multicolumn{1}{c|}{}                          & \multicolumn{1}{c|}{188.09M$^\dagger$}  & \multicolumn{1}{c|}{14.64G} & \multicolumn{1}{c|}{\textbf{5.73} \textbf{\color{ForestGreen}($\times$1.11)}$^\ddagger$}                                                              & 9.39  \\ \hline
        \multicolumn{1}{c|}{OMS-DPM~\cite{liu2023OMS-DPM}}          & \multicolumn{1}{c|}{Searched}                  & \multicolumn{1}{c|}{-}        & \multicolumn{1}{c|}{-}      & \multicolumn{1}{c|}{2.56}                                                              & 11.10 \\
        \multicolumn{1}{c|}{DiffPruning (50\%)} & \multicolumn{1}{c|}{DDIM-200}                  & \multicolumn{1}{c|}{112.6M$^\dagger$}   & \multicolumn{1}{c|}{10.48G} & \multicolumn{1}{c|}{3.15}                                                              & 10.22 \\
        \multicolumn{1}{c|}{DiffPruning (50\%)} & \multicolumn{1}{c|}{DDIM-100}                  & \multicolumn{1}{c|}{112.6M$^\dagger$}   & \multicolumn{1}{c|}{10.48G} & \multicolumn{1}{c|}{\textbf{6.28} \textbf{\color{ForestGreen}($\times$1.21)}$^\ddagger$}                                                       & 10.89 \\ \hline
        \multicolumn{1}{c|}{OMS-DPM~\cite{liu2023OMS-DPM}}          & \multicolumn{1}{c|}{Searched}                  & \multicolumn{1}{c|}{-}        & \multicolumn{1}{c|}{-}      & \multicolumn{1}{c|}{6.4}                                                               & 13.7  \\
        \multicolumn{1}{c|}{DiffPruning (30\%)} & \multicolumn{1}{c|}{DDIM-100}                  & \multicolumn{1}{c|}{36.9M$^\dagger$} & \multicolumn{1}{c|}{6.35G}  & \multicolumn{1}{c|}{\textbf{6.87} \textbf{\color{ForestGreen}($\times$1.32)}$^\ddagger$}                                                              & 11.39 \\ \hline
        \end{tabular}
       }
    \caption{}
    \label{lsun_church_results}
    \end{subtable}
    \hspace{0em}
    \begin{subtable}{0.48\linewidth}
        
        \resizebox{\linewidth}{!}{
            \begin{tabular}{ccccc}
            \hline
            \multicolumn{5}{c}{Class-Conditional ImageNet ($256\times256$)}                                                                                                                                                 \\ \hline
            \multicolumn{1}{c|}{}                 & \multicolumn{2}{c|}{Complexity}                             & \multicolumn{2}{c}{Performance}                                                                \\ \hline
            \multicolumn{1}{c|}{Model}            & \multicolumn{1}{c|}{Params}  & \multicolumn{1}{c|}{MACs}    & \multicolumn{1}{c|}{\begin{tabular}[c]{@{}c@{}}Throughput $(\uparrow)$\\ (Sample/Sec)\end{tabular}} & FID $(\downarrow)$  \\ \hline
            \multicolumn{1}{c|}{ADM~\cite{dhariwal2021DMsBeatGans}}              & \multicolumn{1}{c|}{607.9M}  & \multicolumn{1}{c|}{1186.4G} & \multicolumn{1}{c|}{0.07}                                                              & 4.59  \\
            \multicolumn{1}{c|}{LDM~\cite{rombach2022LDM}}              & \multicolumn{1}{c|}{400.82M} & \multicolumn{1}{c|}{108.78G} & \multicolumn{1}{c|}{0.32}                                                              & 3.60  \\ \hline
            \multicolumn{1}{c|}{DiffPruning (70\%)}     & \multicolumn{1}{c|}{250.79M$^\dagger$} & \multicolumn{1}{c|}{76.24G}  & \multicolumn{1}{c|}{\textbf{0.43}~\textbf{\color{ForestGreen}($\times$1.34)}$^\ddagger$}                                                              & 8.03  \\ \hline
            \multicolumn{1}{c|}{LDM 50\% Scratch~\cite{fang2023StructuralPruningforDMs}} & \multicolumn{1}{c|}{189.43M} & \multicolumn{1}{c|}{\tikz[baseline=(current bounding box.center), shadowed/.style={drop shadow={shadow xshift=0ex, shadow yshift=0ex, opacity=0.3}}]{
        \node[shadowed] {52.71G};
    }}  & \multicolumn{1}{c|}{-}                                                                 & 51.45 \\
            \multicolumn{1}{c|}{Taylor~\cite{fang2023StructuralPruningforDMs}}   & \multicolumn{1}{c|}{189.43M} & \multicolumn{1}{c|}{\tikz[baseline=(current bounding box.center), shadowed/.style={drop shadow={shadow xshift=0ex, shadow yshift=0ex, opacity=0.3}}]{
        \node[shadowed] {52.71G};
    }}  & \multicolumn{1}{c|}{-}                                                                 & 11.18 \\
            \multicolumn{1}{c|}{SP~\cite{fang2023StructuralPruningforDMs}}               & \multicolumn{1}{c|}{189.43M} & \multicolumn{1}{c|}{\tikz[baseline=(current bounding box.center), shadowed/.style={drop shadow={shadow xshift=0ex, shadow yshift=0ex, opacity=0.3}}]{
        \node[shadowed] {52.71G};
    }}  & \multicolumn{1}{c|}{-}                                                                 & 9.16  \\
            \multicolumn{1}{c|}{DiffPruning (50\%)}     & \multicolumn{1}{c|}{161.06M$^\dagger$} & \multicolumn{1}{c|}{54.32G}  & \multicolumn{1}{c|}{\textbf{0.56}~\textbf{\color{ForestGreen}($\times$1.75)}$^\ddagger$}                                                              & 8.45  \\ \hline
            \multicolumn{1}{c|}{DiffPruning (30\%)}     & \multicolumn{1}{c|}{79.82M$^\dagger$}  & \multicolumn{1}{c|}{32.71G}  & \multicolumn{1}{c|}{\textbf{0.92}~\textbf{\color{ForestGreen}($\times$2.87)}$^\ddagger$}                                                              & 13.18 \\ \hline
            \end{tabular}
            }
        \caption{}
        \label{imagenet_results}
    \end{subtable}
    \label{all_results}
% \vspace{-28pt}
\end{table*}

%%%%%%%%%%%%%%%%%%%%%%%%%%%%%%%%%%%%%%%%%%%%%%%%%%%%%%%%%%%%%%%%%%%%%%%%%%%%%%%%%%%%%%%%%%%%%%%%%%%%%%%%%%%%%%%%%%%%%%%%%%%%%%%%%%%%%%%%%%%%%%%%%%%%%%%%%%%%%%%%%%%%%%%%%%%%%%%%%%%%%%%%%%%%%%%%%%%%%%%%%%%%%%%%%%%%%%%%%%%%%%
% LSUN-Church

\noindent\textbf{LSUN-Church.}~The architecture MACs of the LDM~\cite{rombach2022LDM} for LSUN-Church is smaller than LDM models for other datasets as its encoder reduces the spatial dimension by 8 \textit{vs.}~4 for others.~Thus, pruning the LDM for LSUN-Church is more challenging than other ones, and~Tab.~\ref{lsun_church_results} summarizes the results.~First, we can observe that DiffPruning $70\%$ achieves drastically better FID than the DDPM model pruned by SP~\cite{fang2023StructuralPruningforDMs} while having almost $9.5\times$ fewer MACs.~Remarkably, DiffPruning $70\%$ achieves better FID than the full DDPM model, illustrating that LDMs have better computation-performance frontier than pixel-space DPMs.~Second, DiffPruning $50\%$ and $30\%$ models can achieve both higher throughput and better FID while requiring more than $7\times$ less training iterations than OMS-DPM~\cite{liu2023OMS-DPM} (details in supplementary).~DiffPruning $50\%$ with the 100-step DDIM~\cite{song2021DDIM} sampler has $2.45\times$ higher throughput (6.28~\textit{vs.}~2.56) than OMS-DPM with a lower FID.~Also, DiffPruning $50\%$ with 200 steps DDIM sampler still has a higher throughput and better FID than OMS-DPM.~Notably, DiffPruning~$50\%$ with 200 steps DDIM sampler can outperform the full DDPM model ($10.22$~\emph{vs.}~$10.58$ FID) while having $4.25\times$ faster sampling throughput.~Finally, DiffPruning $30\%$ has higher throughput ($6.87$~\emph{vs.}~$6.4$ FID) while outperforming OMS-DPM's model by 2.31 FID.~In summary, the comparison results with OMS-DPM demonstrate the benefit of our elastic fine-tuning for our experts that enables our method to gather a model zoo without requiring training several models from scratch, thereby obtaining a higher-performing mixture of efficient experts with much lower training iterations than OMS-DPM.

\noindent\textbf{FFHQ.}~Tab.~\ref{ffhq_results} shows the results on FFHQ.~DiffPruning 70\%/50\% models can achieve close FID scores to LDM~\cite{rombach2022LDM} while enjoying 33\%/81\% throughput speed-ups. In the extreme case of 30\% MACs budget, DiffPruning secures $2.87\times$ speed-up while having a 1.13 worse FID score than LDM, which shows that it can successfully prune the LDM for small-scale datasets like FFHQ as well.

\begin{figure*}[t!]
  \centering
  \includegraphics[scale=0.215]{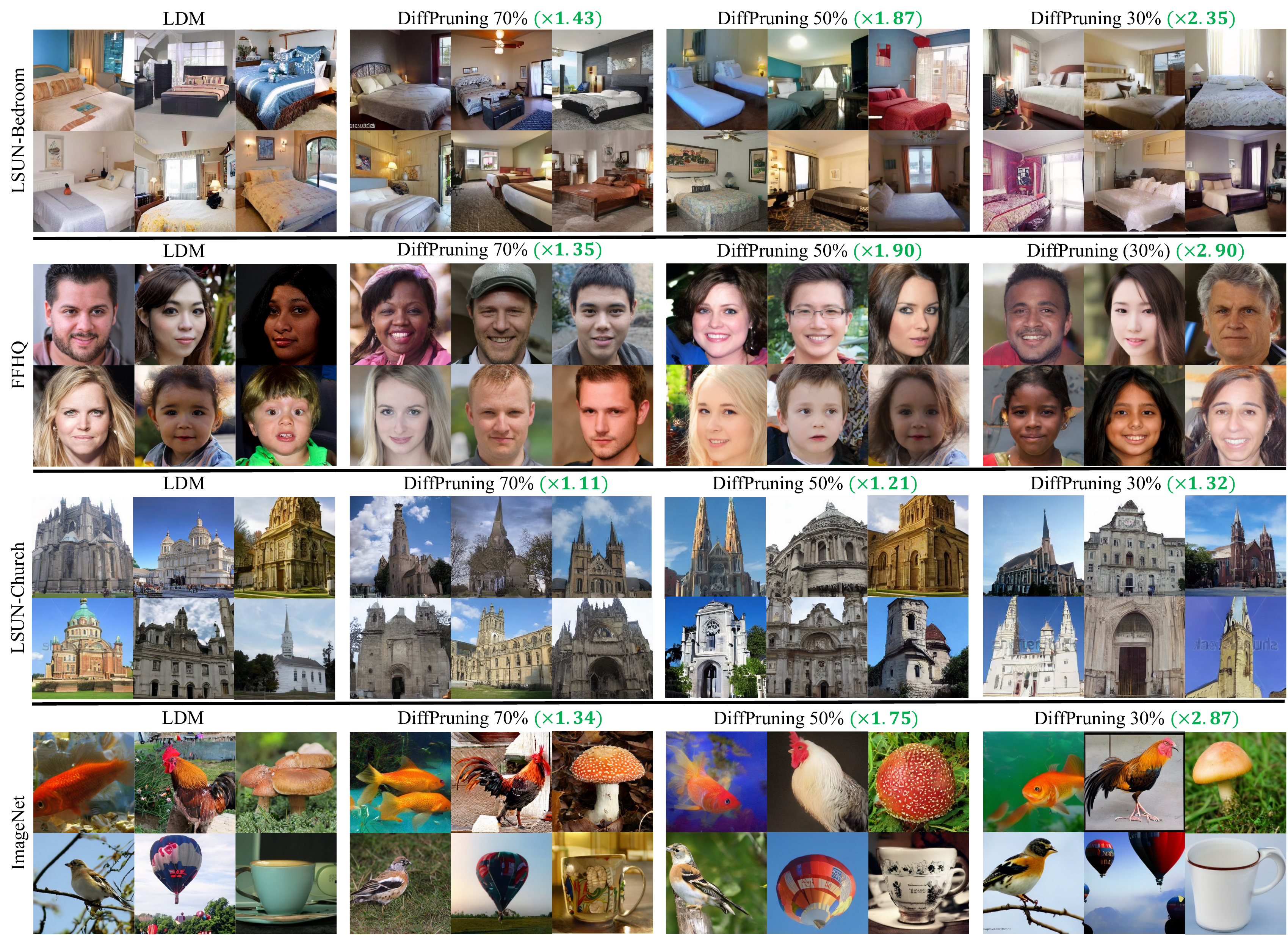}
  \caption{Samples from the LDM~\cite{rombach2022LDM} model and our pruned mixture of experts for different MACs budgets.~The green numbers show the relative sampling throughput speed-up of our pruned models compared to LDM on an NVIDIA A100 GPU.}
  \label{samples_of_models}
  % \vspace{-25pt}
\end{figure*}

\noindent\textbf{Class-Conditional ImageNet.}~Tab.~\ref{imagenet_results} summarizes results for ImageNet.~DiffPruning 50\% model can achieve 0.71 better FID score than SP~\cite{fang2023StructuralPruningforDMs}.~The reported MACs values by SP are not directly comparable to ours, and we elaborate on the reason in supplementary.~In addition, DiffPruning 70\%/30\% obtain 1.34$\times$/2.87$\times$ increase in throughput compared to LDM.~Thus, DiffPruning can effectively prune conditional LDMs as well.~In summary, our experimental results demonstrate that our method can effectively prune both unconditional and conditional LDM models for datasets with various scales. We provide samples generated by the original LDM and our pruned mixture of experts with different budgets in Fig.~\ref{samples_of_models}.

\subsection{Ablation Study}
We conduct an ablation study to explore the contribution of each component of our method to its final performance. We implement a Baseline that uses naive parameterizations $\mathbf{v}$ = $\text{sigmoid}(-(\beta_v + n)/\tau)$ (Eq.~\ref{width-gumbell}) and $\mathbf{u}$ = $\text{sigmoid}(-(\beta_u + n)/\tau)$ (Eq.~\ref{depth-gumbell}) for pruning \underline{a single model}.~Then, we add the mixture of experts, our ERA model, elastic depth fine-tuning, and elastic width fine-tuning one at a time for pruning the model.~Tab.~\ref{ablation_results} summarizes the results.~First, we can observe that employing the mixture of experts improves both the sample quality FID score (which is aligned with the prior works~\cite{balaji2022ediffiMOE,feng2023ernieMOE}) and the inference throughput of the pruned model.~This result quantitatively justifies our design choice for clustering the denoising timesteps into intervals and using a specialized model for each of them.~Employing our ERA model yields a faster model than the naive parameterization.~The reason may be that the naive parameterization cannot properly model the complex interactions between experts and between different layers within an expert.~Noticeably, employing each component of our method improves \textit{both dimensions} of sampling throughput and sample quality such that our method can obtain a 1.28$\times$ faster model (throughput 3.75 \textit{vs.} 2.92) with 3.59 better FID than the naive Baseline.~In summary, our ablation experiments verify our design choices for DiffPruning.

\begin{table}[hbt!]
\centering
\caption{Ablation results of our proposed method for pruning the LDM model~\cite{rombach2022LDM} for LSUN-Bedroom to 50$\%$ MACs budget.}
\resizebox{0.7\linewidth}{!}{
\begin{tabular}{c|c|c|c|c}
\hline
Model                  & Sampler                   & MACs                    & \begin{tabular}[c]{@{}c@{}}Throughput$(\uparrow)$\\ (Sample/Sec)\end{tabular} & FID $(\downarrow)$   \\ \hline
Baseline               & \multirow{6}{*}{DDIM-100} & \multirow{5}{*}{50.69G} & 2.92                                                              & 10.32 \\
+ Mixture of Experts   &                           &                         & 3.05                                                              & 9.65  \\
+ Expert Routing Agent &                           &                         & 3.25                                                              & 8.53  \\
+ Elastic Depth        &                           &                         & 3.61                                                              & 8.03  \\
+ Elastic Width (Ours) &                           &                         & 3.75                                                              & 6.73  \\ \cline{1-1} \cline{3-5} 
LDM~\cite{rombach2022LDM}                    &                           & 101.32G                 & 2.01                                                              & 4.39  \\ \hline
\end{tabular}}
\label{ablation_results}
% \vspace{-10pt}
\end{table}

\section{Conclusions}
% \vspace{-3pt}
We introduce a novel approach for pruning an LDM model into a mixture of efficient experts in which each expert performs the denoising task on an interval of the denoising path.~We employ the model's denoising timesteps' alignment scores to cluster them into several intervals and empirically show that the optimal cluster assignments are different for distinct datasets.~Thus, using static clustering schemes is sub-optimal.~We propose to fine-tune the pre-trained LDM on each cluster interval with elastic dimensions to obtain our interval experts.~By doing so, each expert contains an implicit model zoo within itself for its corresponding interval.~Finally, we develop a new pruning scheme in which our Expert Routing Agent (ERA) learns to prune the elastically trained experts together in an end-to-end manner.~Thus, our ERA automatically allocates the compute budget between experts. Our experimental results validate our method's effectiveness, and our ablation studies show that our design choices improve both dimensions of the pruned model's throughput and its sample quality.

\section*{Acknowledgments}

Alireza Ganjdanesh and Heng Huang were partially supported by NSF IIS 2347592, 2347604, 2348159, 2348169, DBI 2405416, CCF 2348306, CNS 2347617.

\bibliographystyle{splncs04}
\bibliography{main}

\begin{thebibliography}{10}
\providecommand{\url}[1]{\texttt{#1}}
\providecommand{\urlprefix}{URL }
\providecommand{\doi}[1]{https://doi.org/#1}

\bibitem{balaji2022ediffiMOE}
Balaji, Y., Nah, S., Huang, X., Vahdat, A., Song, J., Kreis, K., Aittala, M., Aila, T., Laine, S., Catanzaro, B., et~al.: ediffi: Text-to-image diffusion models with an ensemble of expert denoisers. arXiv preprint arXiv:2211.01324  (2022)

\bibitem{bao2022AnalyticDPM}
Bao, F., Li, C., Zhu, J., Zhang, B.: Analytic-{DPM}: an analytic estimate of the optimal reverse variance in diffusion probabilistic models. In: International Conference on Learning Representations (2022), \url{https://openreview.net/forum?id=0xiJLKH-ufZ}

\bibitem{bengio2013STE}
Bengio, Y., L{\'e}onard, N., Courville, A.: Estimating or propagating gradients through stochastic neurons for conditional computation. arXiv preprint arXiv:1308.3432  (2013)

\bibitem{betker2023dalle3}
Betker, J., Goh, G., Jing, L., Brooks, T., Wang, J., Li, L., Ouyang, L., Zhuang, J., Lee, J., Guo, Y., et~al.: Improving image generation with better captions. Computer Science. https://cdn. openai. com/papers/dall-e-3. pdf  \textbf{2}(3), ~8 (2023)

\bibitem{Cai2020OFA}
Cai, H., Gan, C., Wang, T., Zhang, Z., Han, S.: Once-for-all: Train one network and specialize it for efficient deployment. In: International Conference on Learning Representations (2020), \url{https://openreview.net/forum?id=HylxE1HKwS}

\bibitem{Cai2020OnceForAll}
Cai, H., Gan, C., Wang, T., Zhang, Z., Han, S.: Once-for-all: Train one network and specialize it for efficient deployment. In: International Conference on Learning Representations (2020), \url{https://openreview.net/forum?id=HylxE1HKwS}

\bibitem{cheng2023SurveyDNNPruning}
Cheng, H., Zhang, M., Shi, J.Q.: A survey on deep neural network pruning-taxonomy, comparison, analysis, and recommendations. arXiv preprint arXiv:2308.06767  (2023)

\bibitem{ChoGRU}
Cho, K., van Merrienboer, B., G{\"{u}}l{\c{c}}ehre, {\c{C}}., Bahdanau, D., Bougares, F., Schwenk, H., Bengio, Y.: Learning phrase representations using {RNN} encoder-decoder for statistical machine translation. In: Proceedings of the 2014 Conference on Empirical Methods in Natural Language Processing, {EMNLP} 2014, October 25-29, 2014, Doha, Qatar, {A} meeting of SIGDAT, a Special Interest Group of the {ACL}. {ACL} (2014). \doi{10.3115/v1/d14-1179}, \url{https://doi.org/10.3115/v1/d14-1179}

\bibitem{choi2022PerceptionPrioritized}
Choi, J., Lee, J., Shin, C., Kim, S., Kim, H., Yoon, S.: Perception prioritized training of diffusion models. In: Proceedings of the IEEE/CVF Conference on Computer Vision and Pattern Recognition. pp. 11472--11481 (2022)

\bibitem{dhariwal2021DMsBeatGans}
Dhariwal, P., Nichol, A.: Diffusion models beat gans on image synthesis. Advances in neural information processing systems  \textbf{34},  8780--8794 (2021)

\bibitem{esser2021vqgan}
Esser, P., Rombach, R., Ommer, B.: Taming transformers for high-resolution image synthesis. In: Proceedings of the IEEE/CVF conference on computer vision and pattern recognition. pp. 12873--12883 (2021)

\bibitem{fang2023StructuralPruningforDMs}
Fang, G., Ma, X., Wang, X.: Structural pruning for diffusion models. In: Advances in Neural Information Processing Systems (2023)

\bibitem{feng2023ernieMOE}
Feng, Z., Zhang, Z., Yu, X., Fang, Y., Li, L., Chen, X., Lu, Y., Liu, J., Yin, W., Feng, S., et~al.: Ernie-vilg 2.0: Improving text-to-image diffusion model with knowledge-enhanced mixture-of-denoising-experts. In: Proceedings of the IEEE/CVF Conference on Computer Vision and Pattern Recognition. pp. 10135--10145 (2023)

\bibitem{ganjdanesh2024MGGC}
Ganjdanesh, A., Gao, S., Alipanah, H., Huang, H.: Compressing image-to-image translation gans using local density structures on their learned manifold. In: Proceedings of the AAAI Conference on Artificial Intelligence. vol.~38, pp. 12118--12126 (2024)

\bibitem{ganjdanesh2022ISP}
Ganjdanesh, A., Gao, S., Huang, H.: Interpretations steered network pruning via amortized inferred saliency maps. In: European Conference on Computer Vision. pp. 278--296. Springer (2022)

\bibitem{ganjdanesh2023effconv}
Ganjdanesh, A., Gao, S., Huang, H.: Effconv: efficient learning of kernel sizes for convolution layers of cnns. In: Proceedings of the AAAI Conference on Artificial Intelligence. vol.~37, pp. 7604--7612 (2023)

\bibitem{ganjdanesh2024RLAL}
Ganjdanesh, A., Gao, S., Huang, H.: Jointly training and pruning cnns via learnable agent guidance and alignment. In: Proceedings of the IEEE/CVF Conference on Computer Vision and Pattern Recognition. pp. 16058--16069 (2024)

\bibitem{gao2023ImplicitSuperResolution}
Gao, S., Liu, X., Zeng, B., Xu, S., Li, Y., Luo, X., Liu, J., Zhen, X., Zhang, B.: Implicit diffusion models for continuous super-resolution. In: Proceedings of the IEEE/CVF Conference on Computer Vision and Pattern Recognition. pp. 10021--10030 (2023)

\bibitem{go2023NegativeTransfer}
Go, H., Kim, J., Lee, Y., Lee, S., Oh, S., Moon, H., Choi, S.: Addressing negative transfer in diffusion models. In: Thirty-seventh Conference on Neural Information Processing Systems (2023), \url{https://openreview.net/forum?id=3G2ec833mW}

\bibitem{goodfellow2014GAN}
Goodfellow, I., Pouget-Abadie, J., Mirza, M., Xu, B., Warde-Farley, D., Ozair, S., Courville, A., Bengio, Y.: Generative adversarial nets. Advances in neural information processing systems  \textbf{27} (2014)

\bibitem{gumbel1954GumbellDistribution}
Gumbel, E.J.: Statistical theory of extreme values and some practical applications: a series of lectures, vol.~33. US Government Printing Office (1954)

\bibitem{habibian2023clockwork}
Habibian, A., Ghodrati, A., Fathima, N., Sautiere, G., Garrepalli, R., Porikli, F., Petersen, J.: Clockwork diffusion: Efficient generation with model-step distillation. arXiv preprint arXiv:2312.08128  (2023)

\bibitem{han2015WeightPruning}
Han, S., Pool, J., Tran, J., Dally, W.: Learning both weights and connections for efficient neural network. Advances in neural information processing systems  \textbf{28} (2015)

\bibitem{he2016ResNet}
He, K., Zhang, X., Ren, S., Sun, J.: Deep residual learning for image recognition. In: Proceedings of the IEEE conference on computer vision and pattern recognition. pp. 770--778 (2016)

\bibitem{YangSurveyStructuredPruning}
He, Y., Xiao, L.: Structured pruning for deep convolutional neural networks: A survey. IEEE Transactions on Pattern Analysis and Machine Intelligence pp. 1--20 (2023). \doi{10.1109/TPAMI.2023.3334614}

\bibitem{he2018AutomaticModelCompression}
He, Y., Lin, J., Liu, Z., Wang, H., Li, L.J., Han, S.: Amc: Automl for model compression and acceleration on mobile devices. In: Proceedings of the European conference on computer vision (ECCV). pp. 784--800 (2018)

\bibitem{ho2022imagen_video}
Ho, J., Chan, W., Saharia, C., Whang, J., Gao, R., Gritsenko, A., Kingma, D.P., Poole, B., Norouzi, M., Fleet, D.J., et~al.: Imagen video: High definition video generation with diffusion models. arXiv preprint arXiv:2210.02303  (2022)

\bibitem{ho2020ddpm}
Ho, J., Jain, A., Abbeel, P.: Denoising diffusion probabilistic models. Advances in neural information processing systems  \textbf{33},  6840--6851 (2020)

\bibitem{hou2020DynaBERT}
Hou, L., Huang, Z., Shang, L., Jiang, X., Chen, X., Liu, Q.: Dynabert: Dynamic bert with adaptive width and depth. Advances in Neural Information Processing Systems  \textbf{33},  9782--9793 (2020)

\bibitem{jang2017GumbelSoftmax}
Jang, E., Gu, S., Poole, B.: Categorical reparameterization with gumbel-softmax. In: International Conference on Learning Representations (2017), \url{https://openreview.net/forum?id=rkE3y85ee}

\bibitem{karras2019StyleGAN_FFHQ}
Karras, T., Laine, S., Aila, T.: A style-based generator architecture for generative adversarial networks. In: Proceedings of the IEEE/CVF conference on computer vision and pattern recognition. pp. 4401--4410 (2019)

\bibitem{kingma2021variationalDMs}
Kingma, D., Salimans, T., Poole, B., Ho, J.: Variational diffusion models. Advances in neural information processing systems  \textbf{34},  21696--21707 (2021)

\bibitem{KingmaVAE}
Kingma, D.P., Welling, M.: Auto-encoding variational bayes. In: Bengio, Y., LeCun, Y. (eds.) 2nd International Conference on Learning Representations, {ICLR} 2014, Banff, AB, Canada, April 14-16, 2014, Conference Track Proceedings (2014), \url{http://arxiv.org/abs/1312.6114}

\bibitem{lee2023MEME}
Lee, Y., Kim, J.Y., Go, H., Jeong, M., Oh, S., Choi, S.: Multi-architecture multi-expert diffusion models. arXiv preprint arXiv:2306.04990  (2023)

\bibitem{li2017L1NormPruning}
Li, H., Kadav, A., Durdanovic, I., Samet, H., Graf, H.P.: Pruning filters for efficient convnets. In: International Conference on Learning Representations (2017), \url{https://openreview.net/forum?id=rJqFGTslg}

\bibitem{li2017PruningFilters}
Li, H., Kadav, A., Durdanovic, I., Samet, H., Graf, H.P.: Pruning filters for efficient convnets. In: International Conference on Learning Representations (2017), \url{https://openreview.net/forum?id=rJqFGTslg}

\bibitem{liu2023OMS-DPM}
Liu, E., Ning, X., Lin, Z., Yang, H., Wang, Y.: Oms-dpm: Optimizing the model schedule for diffusion probabilistic models. In: International Conference on Machine Learning. pp. 21915--21936. PMLR (2023)

\bibitem{liu2018DARTS}
Liu, H., Simonyan, K., Yang, Y.: {DARTS}: Differentiable architecture search. In: International Conference on Learning Representations (2019), \url{https://openreview.net/forum?id=S1eYHoC5FX}

\bibitem{liu2022PseudoNumerical}
Liu, L., Ren, Y., Lin, Z., Zhao, Z.: Pseudo numerical methods for diffusion models on manifolds. In: International Conference on Learning Representations (2022), \url{https://openreview.net/forum?id=PlKWVd2yBkY}

\bibitem{loshchilov2018AdamW}
Loshchilov, I., Hutter, F.: Decoupled weight decay regularization. In: International Conference on Learning Representations (2019), \url{https://openreview.net/forum?id=Bkg6RiCqY7}

\bibitem{lu2022DPMSolver}
Lu, C., Zhou, Y., Bao, F., Chen, J., Li, C., Zhu, J.: Dpm-solver: A fast ode solver for diffusion probabilistic model sampling in around 10 steps. Advances in Neural Information Processing Systems  \textbf{35},  5775--5787 (2022)

\bibitem{lu2023dpmsolver++}
Lu, C., Zhou, Y., Bao, F., Chen, J., Li, C., Zhu, J.: {DPM}-solver++: Fast solver for guided sampling of diffusion probabilistic models (2023), \url{https://openreview.net/forum?id=4vGwQqviud5}

\bibitem{ma2023DeepCache}
Ma, X., Fang, G., Wang, X.: Deepcache: Accelerating diffusion models for free. arXiv preprint arXiv:2312.00858  (2023)

\bibitem{maddison2017ConcreteDistribution}
Maddison, C.J., Mnih, A., Teh, Y.W.: The concrete distribution: A continuous relaxation of discrete random variables. In: International Conference on Learning Representations (2017), \url{https://openreview.net/forum?id=S1jE5L5gl}

\bibitem{meng2023distillation}
Meng, C., Rombach, R., Gao, R., Kingma, D., Ermon, S., Ho, J., Salimans, T.: On distillation of guided diffusion models. In: Proceedings of the IEEE/CVF Conference on Computer Vision and Pattern Recognition. pp. 14297--14306 (2023)

\bibitem{Molchanov2019TaylorPruning}
Molchanov, P., Mallya, A., Tyree, S., Frosio, I., Kautz, J.: Importance estimation for neural network pruning. In: Proceedings of the IEEE/CVF Conference on Computer Vision and Pattern Recognition (CVPR) (June 2019)

\bibitem{nichol2021improvedDDPM}
Nichol, A.Q., Dhariwal, P.: Improved denoising diffusion probabilistic models. In: International Conference on Machine Learning. pp. 8162--8171. PMLR (2021)

\bibitem{pan2024TStich}
Pan, Z., Zhuang, B., Huang, D.A., Nie, W., Yu, Z., Xiao, C., Cai, J., Anandkumar, A.: T-stitch: Accelerating sampling in pre-trained diffusion models with trajectory stitching. arXiv preprint arXiv:2402.14167  (2024)

\bibitem{paszke2019pytorch}
Paszke, A., Gross, S., Massa, F., Lerer, A., Bradbury, J., Chanan, G., Killeen, T., Lin, Z., Gimelshein, N., Antiga, L., et~al.: Pytorch: An imperative style, high-performance deep learning library. Advances in neural information processing systems  \textbf{32} (2019)

\bibitem{rombach2022LDM}
Rombach, R., Blattmann, A., Lorenz, D., Esser, P., Ommer, B.: High-resolution image synthesis with latent diffusion models. In: Proceedings of the IEEE/CVF conference on computer vision and pattern recognition. pp. 10684--10695 (2022)

\bibitem{ronneberger2015UNet}
Ronneberger, O., Fischer, P., Brox, T.: U-net: Convolutional networks for biomedical image segmentation. In: Medical Image Computing and Computer-Assisted Intervention--MICCAI 2015: 18th International Conference, Munich, Germany, October 5-9, 2015, Proceedings, Part III 18. pp. 234--241. Springer (2015)

\bibitem{RussakovskyImageNet}
Russakovsky, O., Deng, J., Su, H., Krause, J., Satheesh, S., Ma, S., Huang, Z., Karpathy, A., Khosla, A., Bernstein, M., Berg, A.C., Fei-Fei, L.: {ImageNet Large Scale Visual Recognition Challenge}. International Journal of Computer Vision (IJCV)  \textbf{115}(3),  211--252 (2015). \doi{10.1007/s11263-015-0816-y}

\bibitem{saharia2022SR3}
Saharia, C., Ho, J., Chan, W., Salimans, T., Fleet, D.J., Norouzi, M.: Image super-resolution via iterative refinement. IEEE Transactions on Pattern Analysis and Machine Intelligence  \textbf{45}(4),  4713--4726 (2022)

\bibitem{salimans2022ProgressiveDistillation}
Salimans, T., Ho, J.: Progressive distillation for fast sampling of diffusion models. In: International Conference on Learning Representations (2022), \url{https://openreview.net/forum?id=TIdIXIpzhoI}

\bibitem{sohl2015nonequilibrium}
Sohl-Dickstein, J., Weiss, E., Maheswaranathan, N., Ganguli, S.: Deep unsupervised learning using nonequilibrium thermodynamics. In: International conference on machine learning. pp. 2256--2265. PMLR (2015)

\bibitem{song2021DDIM}
Song, J., Meng, C., Ermon, S.: Denoising diffusion implicit models. In: International Conference on Learning Representations (2021), \url{https://openreview.net/forum?id=St1giarCHLP}

\bibitem{song2019-estimating-grads-of-data}
Song, Y., Ermon, S.: Generative modeling by estimating gradients of the data distribution. Advances in neural information processing systems  \textbf{32} (2019)

\bibitem{vahdat2021score}
Vahdat, A., Kreis, K., Kautz, J.: Score-based generative modeling in latent space. Advances in Neural Information Processing Systems  \textbf{34},  11287--11302 (2021)

\bibitem{vaswani2017Attention}
Vaswani, A., Shazeer, N., Parmar, N., Uszkoreit, J., Jones, L., Gomez, A.N., Kaiser, {\L}., Polosukhin, I.: Attention is all you need. Advances in neural information processing systems  \textbf{30} (2017)

\bibitem{watson2022LearningFastSamplers}
Watson, D., Chan, W., Ho, J., Norouzi, M.: Learning fast samplers for diffusion models by differentiating through sample quality. In: International Conference on Learning Representations (2022), \url{https://openreview.net/forum?id=VFBjuF8HEp}

\bibitem{watson2022learning}
Watson, D., Ho, J., Norouzi, M., Chan, W.: Learning to efficiently sample from diffusion probabilistic models (2022), \url{https://openreview.net/forum?id=LOz0xDpw4Y}

\bibitem{white2023NAS1000Papers}
White, C., Safari, M., Sukthanker, R., Ru, B., Elsken, T., Zela, A., Dey, D., Hutter, F.: Neural architecture search: Insights from 1000 papers. arXiv preprint arXiv:2301.08727  (2023)

\bibitem{xu2023RestartSampling}
Xu, Y., Deng, M., Cheng, X., Tian, Y., Liu, Z., Jaakkola, T.: Restart sampling for improving generative processes. arXiv preprint arXiv:2306.14878  (2023)

\bibitem{yang2023DPMsMadeSlim}
Yang, X., Zhou, D., Feng, J., Wang, X.: Diffusion probabilistic model made slim. In: Proceedings of the IEEE/CVF Conference on Computer Vision and Pattern Recognition. pp. 22552--22562 (2023)

\bibitem{yao2021JointDetNAS}
Yao, L., Pi, R., Xu, H., Zhang, W., Li, Z., Zhang, T.: Joint-detnas: Upgrade your detector with nas, pruning and dynamic distillation. In: Proceedings of the IEEE/CVF conference on computer vision and pattern recognition. pp. 10175--10184 (2021)

\bibitem{yu15lsun}
Yu, F., Zhang, Y., Song, S., Seff, A., Xiao, J.: Lsun: Construction of a large-scale image dataset using deep learning with humans in the loop. arXiv preprint arXiv:1506.03365  (2015)

\bibitem{zhang2023TailoredMultiDecoder}
Zhang, H., Lu, Y., Alkhouri, I., Ravishankar, S., Song, D., Qu, Q.: Improving efficiency of diffusion models via multi-stage framework and tailored multi-decoder architectures. arXiv preprint arXiv:2312.09181  (2023)

\bibitem{zhang2023ControlNet}
Zhang, L., Agrawala, M.: Adding conditional control to text-to-image diffusion models. arXiv preprint arXiv:2302.05543  (2023)

\bibitem{zhang2023ExponentialIntegrator}
Zhang, Q., Chen, Y.: Fast sampling of diffusion models with exponential integrator. In: The Eleventh International Conference on Learning Representations (2023), \url{https://openreview.net/forum?id=Loek7hfb46P}

\bibitem{zhang2023gddim}
Zhang, Q., Tao, M., Chen, Y.: g{DDIM}: Generalized denoising diffusion implicit models. In: The Eleventh International Conference on Learning Representations (2023), \url{https://openreview.net/forum?id=1hKE9qjvz-}

\bibitem{zhao2023Uni-ControlNet}
Zhao, S., Chen, D., Chen, Y.C., Bao, J., Hao, S., Yuan, L., Wong, K.Y.K.: Uni-controlnet: All-in-one control to text-to-image diffusion models. arXiv preprint arXiv:2305.16322  (2023)

\bibitem{zhao2023MobileDiffusion}
Zhao, Y., Xu, Y., Xiao, Z., Hou, T.: Mobilediffusion: Subsecond text-to-image generation on mobile devices. arXiv preprint arXiv:2311.16567  (2023)

\bibitem{zheng2023TruncatedDPMs}
Zheng, H., He, P., Chen, W., Zhou, M.: Truncated diffusion probabilistic models and diffusion-based adversarial auto-encoders. In: The Eleventh International Conference on Learning Representations (2023), \url{https://openreview.net/forum?id=HDxgaKk956l}

\bibitem{zoph2017NAS}
Zoph, B., Le, Q.: Neural architecture search with reinforcement learning. In: International Conference on Learning Representations (2017), \url{https://openreview.net/forum?id=r1Ue8Hcxg}

\end{thebibliography}

% \title{Supplementary Materials for \\ Efficient Diffusion Models Through Mixture of Efficient Experts} 
\title{Supplementary Materials for \\ Mixture of Efficient Diffusion Experts Through Automatic Interval and Sub-Network Selection} 

\titlerunning{Supplementary Materials for DiffPruning}

\author{Alireza Ganjdanesh\inst{1}\thanks{Part of this work was done during an internship at Adobe Research.} \and 
Yan Kang\inst{2} \and
Yuchen Liu\inst{2} \and
Richard Zhang\inst{2} \and
Zhe Lin\inst{2} \and
Heng Huang\inst{1}}

\authorrunning{A.~Ganjdanesh et al.}

\institute{Department of Computer Science, University of Maryland College Park \and
Adobe Research\\
\email{\{aliganj,heng\}@umd.edu,~\{yankang,yuliu,rizhang,zlin\}@adobe.com}}

\maketitle

We elaborate on more details about our method, experimental setup, experimental results, and related work in the supplementary materials.~We follow the same notations introduced in the paper.

\section{Experimental Results}
\begin{figure}[hbt!]
\centering
\includegraphics[scale=0.42]{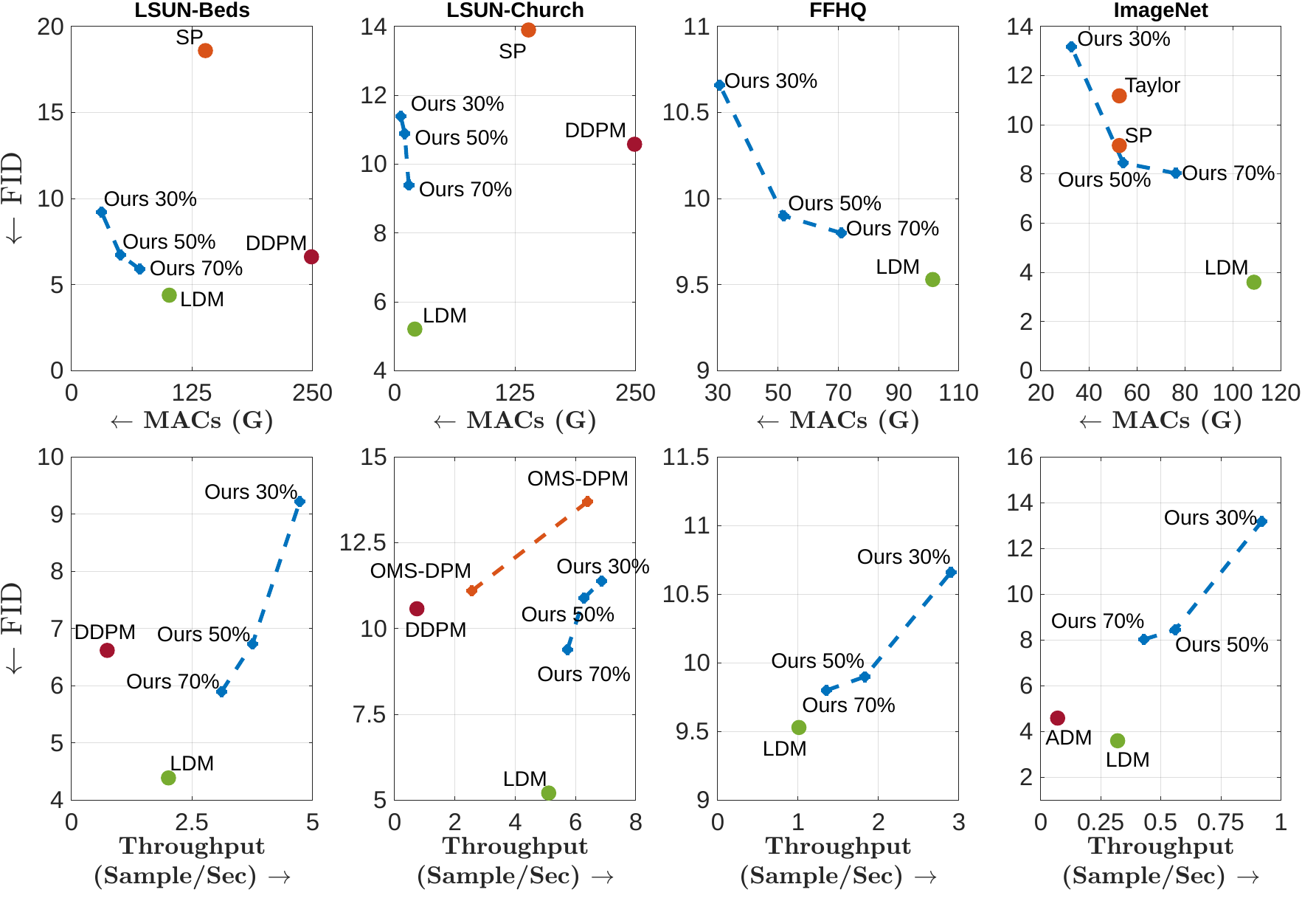}
\caption{Comparison Results of our method \textit{vs.} baselines, SP~\cite{fang2023StructuralPruningforDMs}, OMS-DPM~\cite{liu2023OMS-DPM}, DDPM~\cite{ho2020ddpm}, and LDM~\cite{rombach2022LDM}.~\textbf{First Row:} FID \textit{vs.} MACs curves. \textbf{Second Row:} FID \textit{vs.} Throughput curves. We calculate the Throughput values with an NVIDIA A100 GPU. Higher Throughput and Lower FID and MACs indicate a better performance.}

\vspace{-18pt}
\label{comparison_curves}
\end{figure}

We present the FID~\emph{vs.}~MACs and FID~\emph{vs.}~Throughput of our method and baselines in Fig.~\ref{comparison_curves}. 

\section{Details of Our Method}
\subsection{Clustering Denoising Time-Steps into Intervals}

In this section, we provide details of our method to cluster denoising time-steps of an LDM~\cite{rombach2022LDM} into intervals.~As mentioned in Sec.~(3.3) in the paper, we use two experts in our experiments for computational efficiency.~Thus, we explain our approach for clustering the denoising path into two intervals, but one may extend it to a higher number of intervals as well.

\begin{figure}[hbt!]
  \centering
  \includegraphics[scale=0.5]{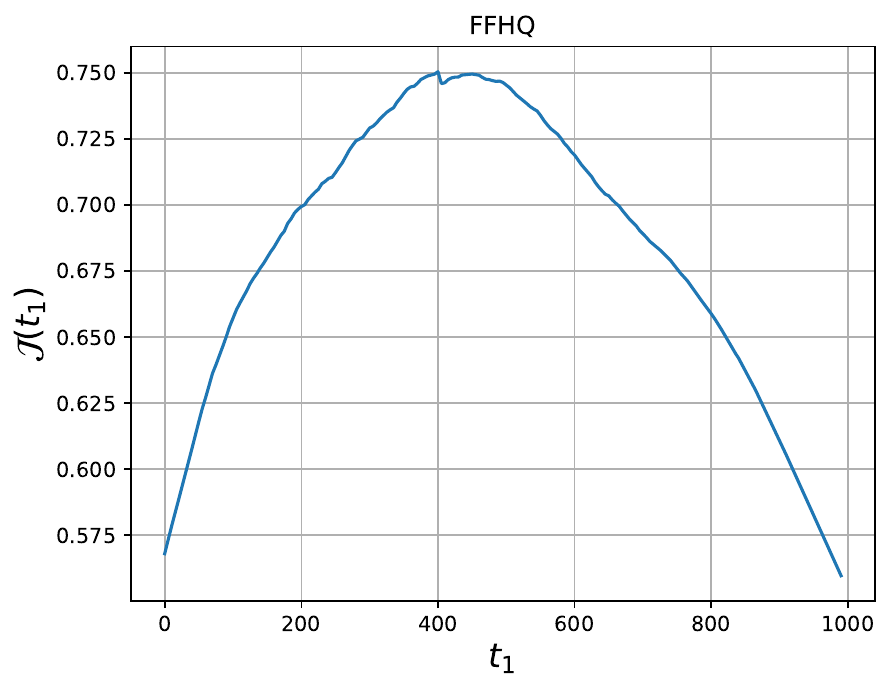}
  \caption{Weighted average $\mathcal{J}(t_1)$~(Eq.~\ref{obj_cutoff}) of the mean of alignment scores in two clusters for the LDM trained on FFHQ.}
\label{ffhq_sim}
\end{figure}

\begin{figure}[hbt!]
  \centering
  \includegraphics[scale=0.5]{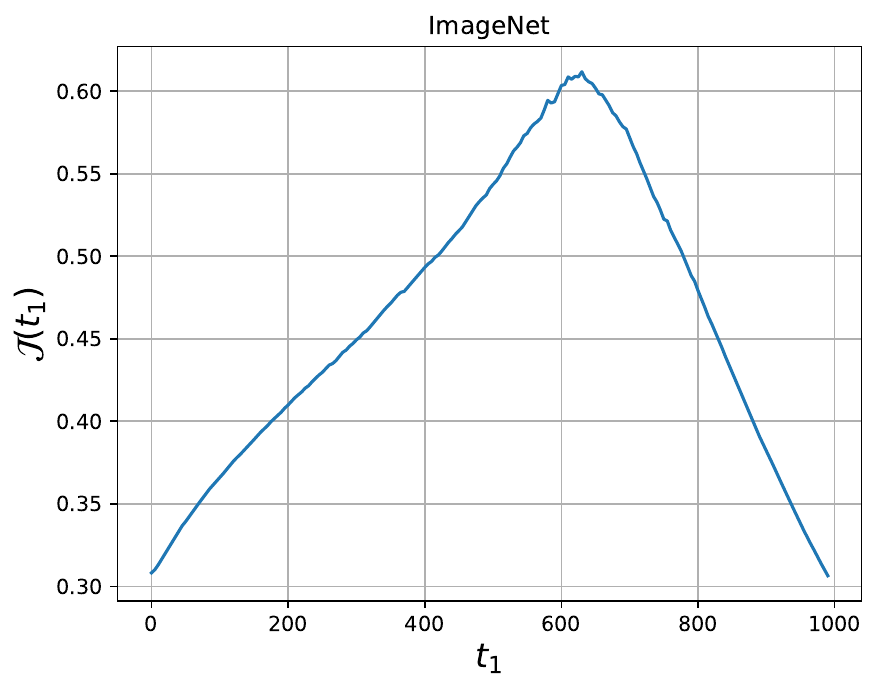}
  \caption{Weighted average $\mathcal{J}(t_1)$~(Eq.~\ref{obj_cutoff}) of the mean of alignment scores in two clusters for the LDM trained on ImageNet.}
\label{imagenet_sim}
\end{figure}

\begin{figure}[hbt!]
  \centering
  \includegraphics[scale=0.5]{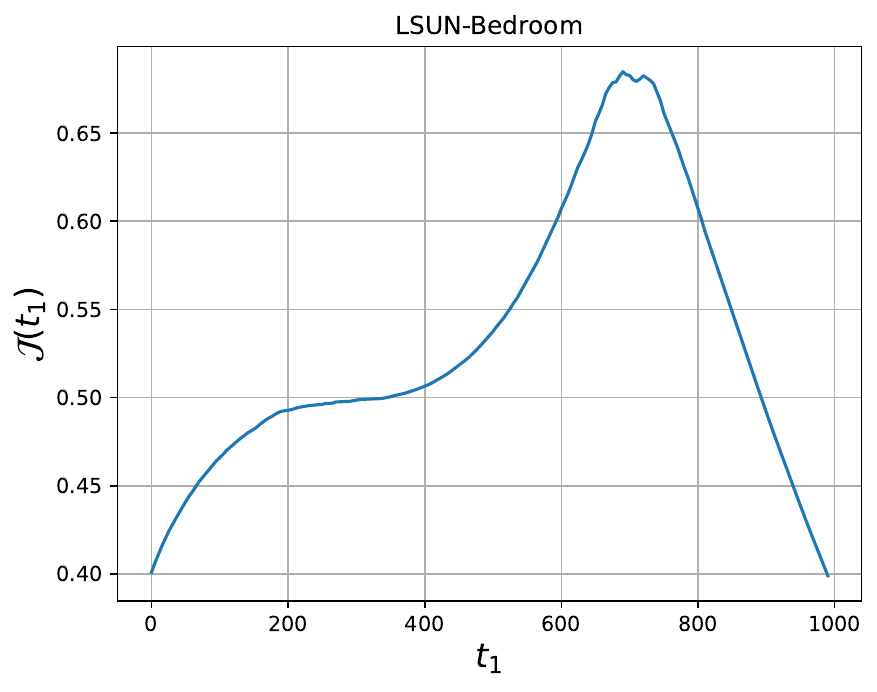}
  \caption{Weighted average $\mathcal{J}(t_1)$~(Eq.~\ref{obj_cutoff}) of the mean of alignment scores in two clusters for the LDM trained on LSUN-Beds.}
\label{lsun_bed_sim}
\end{figure}

\begin{figure}[hbt!]
  \centering
  \includegraphics[scale=0.5]{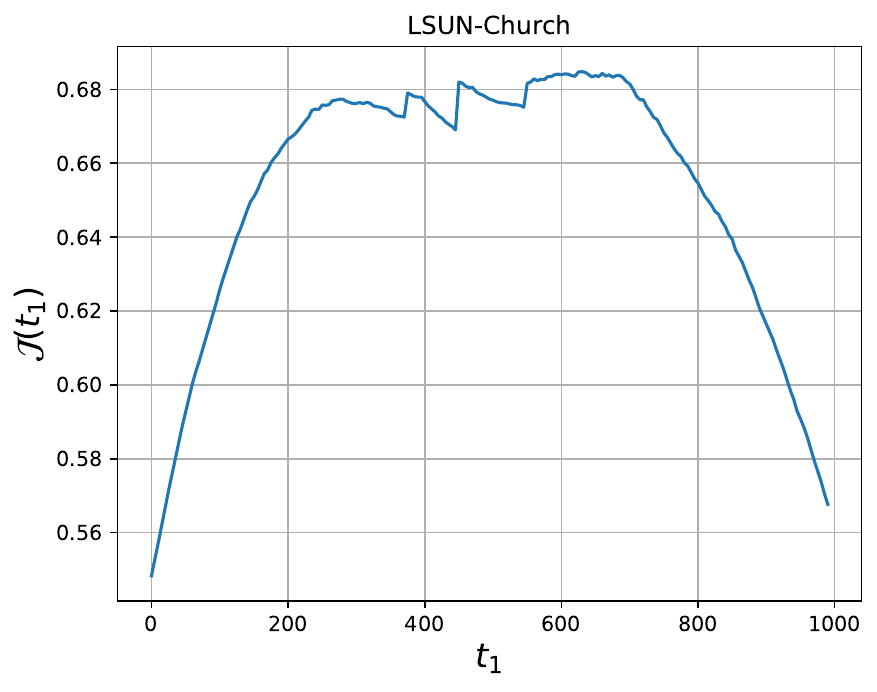}
  \caption{Weighted average $\mathcal{J}(t_1)$~(Eq.~\ref{obj_cutoff}) of the mean of alignment scores in two clusters for the LDM trained on LSUN-Church.} 
  \label{lsun_church_sim}
\end{figure}

Given denoising time-steps $\mathcal{T}=[1, T]$, we divide it into two intervals $\mathcal{I}_1=[T, t_1]$ and $\mathcal{I}_2=[t_1, 1]$.~Now, the main question is how to determine the cut-off time-step $t_1$.~We propose to leverage alignment scores of time-steps to find the optimal $t_1$.~We take the cosine similarity between the gradients of training objectives $\mathcal{L}_t$ and $\mathcal{L}_s$ as the alignment score of the time-steps $t$ and $s$ and denote it with $a_{t, s}$.~We propose to select the cut-off point that maximizes the weighted average of the mean of alignment scores in clusters:

\begin{align}
    \max_{t_1}\mathcal{J}(t_1) &= \sum_{i=1}^{2}[w_i(\sum_{j\in \mathcal{I}_i}\sum_{ k \in \mathcal{I}_i}\frac{a_{j, k}}{|\mathcal{I}_i|^2})]\label{obj_cutoff} \\ 
    w_i &= \frac{|\mathcal{I}_i|}{T}\label{weight}
\end{align}

\noindent where $|\mathcal{I}_i|$ is the number of time-steps in $\mathcal{I}_i$ and $T$ is the total number of denoising time-steps that is usually set to 1000 in practice~\cite{rombach2022LDM,ho2020ddpm,song2021DDIM}.~The Obj~\ref{obj_cutoff} encourages to choose $t_1$ such that the average of alignment scores be high in each cluster while the weights $w_i$ adjust the contribution of each cluster to the objective based on their size.~Our weighting scheme prevents degenerate solutions such as choosing a single time-step as a separate cluster.~Figs~(\ref{ffhq_sim}-\ref{lsun_church_sim}) show the $\mathcal{J}(t_1)$ functions for different datasets.~We choose the cut-off values 400, 625, 700, and 700 for the FFHQ, ImageNet, LSUN-Bedroom, and LSUN-Church models to maximize their $\mathcal{J}(t_1)$ values.~These values result in the $(\mathcal{I}_1, \mathcal{I}_2)$ clusters that we chose in Sec.~(3.3) and Fig.~(3) in the paper.~We believe that one can readily extend our method to the cases with a higher number of experts by optimizing for the cut-off points that maximize similar objectives to $\mathcal{J}$.~Specifically, if one decides to use $C$+$1$ experts (clusters), they should find $C$ cut-off points $t_1$$<$$t_2$$<$$\cdots$$<$$t_{C}$ to optimize the following objective: 

\begin{equation}
    \max_{t_1, t_2,\cdots t_{C}}\mathcal{J}=\sum_{i=1}^{C+1}[w_i(\sum_{j\in \mathcal{I}_i}\sum_{ k \in \mathcal{I}_i}\frac{a_{j, k}}{|\mathcal{I}_i|^2})]
\end{equation}

\noindent with the same definitions as Eqs.~(\ref{obj_cutoff},~\ref{weight}).

Finally, we note that we use 1024 random images to estimate the gradient of each time-step for the models for LSUN-Church~\cite{yu15lsun}, LSUN-Bedroom~\cite{yu15lsun}, and FFHQ~\cite{karras2019StyleGAN_FFHQ}.~We employ 16384 samples to do so on ImageNet~\cite{RussakovskyImageNet}. 

\subsection{Fine-tuning with Elastic Width}\label{elastic_width_finetuning}

\begin{figure}[hbt!]
  \centering
  \includegraphics[scale=0.55]{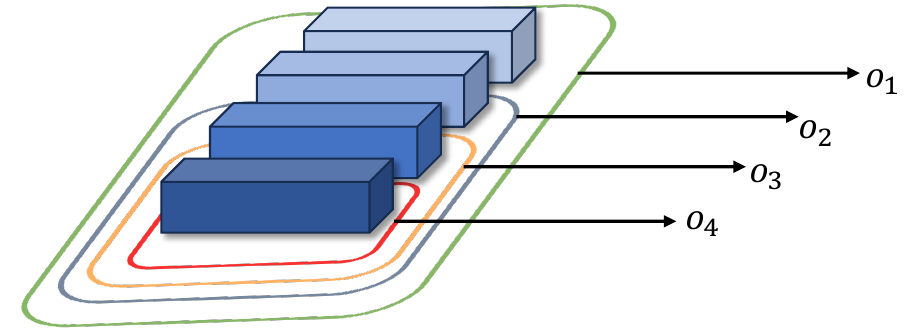}
  \caption{Illustration of our Elastic Width training.~We sort the convolution channels (attention heads) based on their importance ($L_1$ norm) before starting elastic width training.~We drop a random ratio of the least important channels (heads) for convolution layers (attention layers) for each batch of training.~The values $o_{1:4}$ represent different possible dropping ratios for a convolution layer with 4 channels.}
\label{elastic_width}
\end{figure}

We mentioned in Sec.~(3.4) in the paper that we fine-tune the experts with elastic width after training them with elastic depth. Here, width means the channels of the convolution layers in the Residual Blocks~\cite{he2016ResNet} and heads of the attention layers~\cite{vaswani2017Attention} in the attention blocks of the U-Net.~Before starting the elastic width fine-tuning, we sort the channels in the convolution layers in ResBlocks based on their importance, determined by their $L_1$ norm~\cite{Cai2020OFA,li2017L1NormPruning}.~Similarly, we sort the attention heads based on the $L_1$ norm of their projection weights.~Then, in each batch of elastic width fine-tuning, We independently sample a random ratio $r$ ($r\sim \mathcal{U}[0,~1)$) for each convolution layer with $W$ channels (attention layer with $W$ heads) and drop the $\lfloor Wr\rfloor$ least important channels (attention heads) of the layer.~We illustrate our elastic width channel selection for a convolution layer with 4 channels in Fig.~\ref{elastic_width}.~The channels are sorted based on their $L_1$ norm (shown by their color intensity).~The values $o_{1:4}$ represent different possible channel dropping cases for our elastic width training of the convolution layers.~We similarly drop a ratio of least important attention heads.

\subsection{Expert Routing Agent} \label{ERA}
We use a Gated Recurrent Unit (GRU)~\cite{ChoGRU} and dense layers to implement our Expert Routing Agent (ERA).~As mentioned in Sec.~(3.5) in the main text, our ERA predicts architecture vectors ($u^{(i)},~v^{(i)}$) that determine (depth, width) pruning for the expert $E_i$.~We assume that each expert has $D$ depth pruning layers and $L$ layers for width pruning.~We show the detailed architecture in Tab.~\ref{tab-arch}.~We randomly initialize the inputs $z^{(i)}$ and keep them fixed during our pruning process.~The values $C^{(i)}_k$ when $k\in[1,\cdots,L]$ are equal to the widths of the layers.~In addition, $C^{(i)}_{L+1} = D$ as we use the outputs of the $Dense_{L+1}$ layer to calculate the depth architecture vectors $u^{(i)}$.

\begin{table}[hbt!]
\centering
\caption{The architecture of our Expert Routing Agent.~We calculate width architecture vectors $v^{(i)}$ from the outputs $o_k^{(i)}$ ($k\in[1, L]$).~We compute the depth architecture vector $u^{(i)}$ from $o_{L+1}^{(i)}$.~We refer to Sec.~\ref{width_pruning} for detailed formulations.}
    {
        \begin{tabular}{c}
            \toprule
                  Inputs $z^{(i)}=[z^{(i)}_k]$, ($k=1,\cdots, L+1$), ($i = 1,\cdots,N$)\\
            \midrule
                  GRU(128,~256), WeightNorm, ReLU\\
                $\textrm{Dense}_k$(256,~$C^{(i)}_k$), WeightNorm, ($k=1,\cdots, L+1$)\\
             \midrule
                  Outputs ${o}^{(i)}_k$, ($k=1,\cdots, L+1$) \\
                  % $u^{(i)}=[{o}^{(i)}_k]$, ($k=L+1,\cdots, L+D$) \\
                  % $v^{(i)}=[{o}^{(i)}_k]$, ($k=1,\cdots, L$)\\
            \bottomrule
              \\
          \end{tabular}
          }
        \label{tab-arch}
\end{table}

\subsubsection{Formulation of Architecture Vectors}\label{width_pruning}
In this section, we describe our approach to calculate the architecture vectors $(u^{(i)}, v^{(i)})$ from the output vectors $o^{(i)}$ (Tab.~\ref{tab-arch}) of the ERA.

\noindent\textbf{Width Architecture Vectors:}~We design our width pruning method while considering our elastic width fine-tuning scheme.~As mentioned in Sec.~\ref{elastic_width_finetuning} and Fig.~\ref{elastic_width}, we drop a random ratio of the least important convolution channels (attention heads) when training the convolution layers (attention layers) in the elastic width manner.~We call each convolution channel and attention head a `width unit.'~Due to our dropping scheme, the weights of more important width units get trained more than the lower-importance ones in a layer and are robust to removing the lower-importance units.

Accordingly, we embed such a prior into the calculation of our width pruning architecture vectors $v^{(i)}=[v_{l}^{(i)}]_{l=1}^{L}$.~Let's assume that the $l$-th layer of the expert $E_i$ has $W$ width units $[c_w]_{w=1}^{W}$ that are sorted based on their orders, namely $c_1$ is the most important unit, and $c_W$ is the least important one.~As mentioned in Sec.~(3.5.1) of the paper, the calculation from $v^{(i)}$ to $\mathbf{v}^{(i)}$ is called the Gumbel-Sigmoid trick~\cite{maddison2017ConcreteDistribution,jang2017GumbelSoftmax}, which is a differentiable estimation of sampling from a Bernoulli distribution with the Bernoulli parameter of $\texttt{sigmoid(}v^{(i)}\texttt{)}$.~We calculate the vector $v_{l}^{(i)}=[v_{l, w}^{(i)}]_{w=1}^{W}$ from the output vector $o_{l}^{(i)}$ (Tab.~\ref{tab-arch}) such that the Bernoulli parameters $\texttt{sigmoid(}v^{(i)}_{l, w}\texttt{)}$ follow the importance order for the width units $c_w$. By doing so, the probability of preserving the more important width units is higher than low-importance ones. Specifically, we calculate $v_{l}^{(i)}$ as follows:

\begin{align}
    y_{l}^{(i)} &= \text{Softmax}(o_{l}^{(i)}) \\
    p_{l}^{(i)} &= \text{cumsum}(y_{l}^{(i)}) \\
    v_{l}^{(i)} &= \text{inverse-sigmoid}(p_{l}^{(i)} - \epsilon) \label{v_calculation}
\end{align}

\noindent In other words, first, we calculate the Softmax of the output logits vector $o_{l}^{(i)}$.~Then, we take the cumulative summation of the elements of $y_{l}^{(i)}$ as $p_{l}^{(i)}$ such that $p_{l, e}^{(i)} = \sum_{w=e+1}^{W}y_{l, w}^{(i)}$.~Thus, $p_{l,1}^{(i)} > p_{l,2}^{(i)} > \cdots > p_{l,W}^{(i)}$.~Finally, we calculate the inverse sigmoid function for the elements of the probability vector $p_{l}^{(i)}$ to obtain the vector $v_{l}^{(i)}$~(the small constant $\epsilon$ ensures numerical stability of the inverse sigmoid function).~Doing so ensures that the ERA will preserve the more important width units with a higher probability than the low-importance ones.

\begin{figure}[hbt!]
\centering
\includegraphics[scale=0.4]{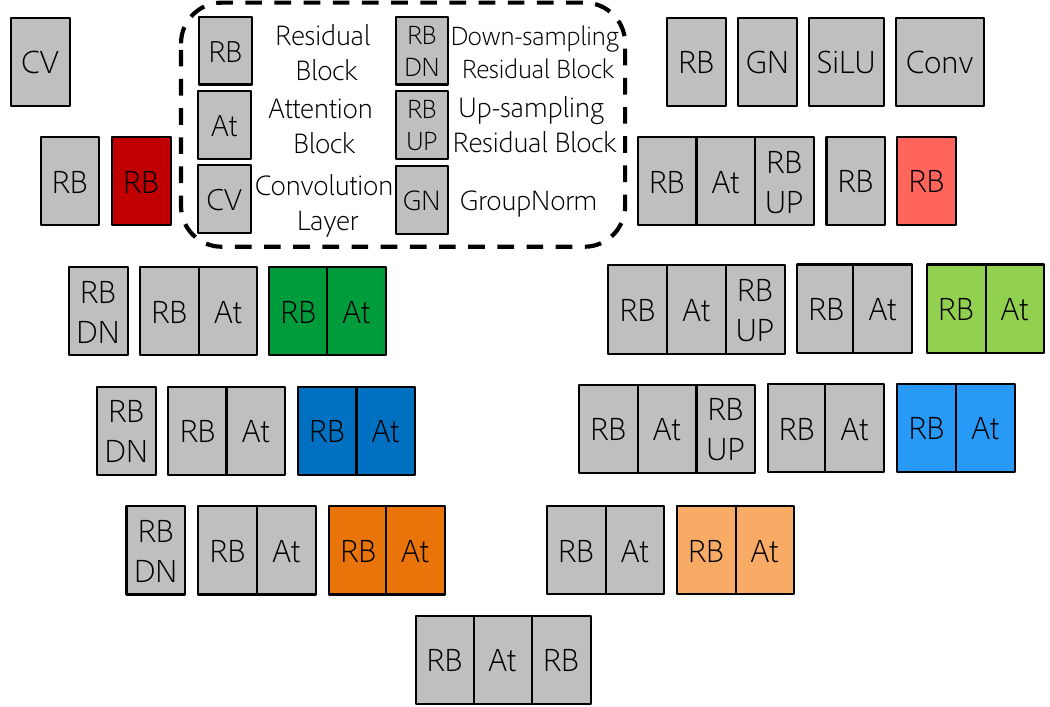}
\caption{U-Net architecture of the LDM~\cite{rombach2022LDM}.}
\label{ldm_unet}
\end{figure}

\noindent\textbf{Depth Architecture Vectors:}~We calculate the depth architecture vectors $u^{(i)}$ similar to the scheme for $v^{(i)}_l$:  

\begin{align}
    y_{L+1}^{(i)} &= \text{Softmax}(o_{L+1}^{(i)}) \\
    p_{L+1}^{(i)} &= \text{cumsum}(y_{L+1}^{(i)}) \\
    u^{(i)} &= \text{inverse-sigmoid}(p_{L+1}^{(i)} - \epsilon) \label{u_calculation}
\end{align}

For the depth layers, we empirically found that removing the shallower depth layers results in a more severe increase in training loss values as well as degradation in sample quality.~Similarly we observed that for the depth blocks in the same stage of the U-Net, the depth block in the decoder branch is more crucial to the model's performance than the one in the encoder branch.~Thus, we rank the depth blocks based on 1) their stage and 2) the branch of the U-Net that the block belongs to.~For instance, we rank the depth pruning blocks in Fig~\ref{ldm_unet} as: $r$ = [decoder red block, encoder red block, decoder green block, encoder green block, decoder blue block, encoder blue block, decoder orange block, encoder orange block].~We then apply the elements of the soft relaxed depth pruning architecture vector $\mathbf{u}^{(i)}$ calculated from $u^{(i)}$ (Eq. 9 in the main text) with the same order as the ranking $r$ to the depth blocks.~By doing so, the probabilities that our method preserve the depth pruning blocks will have the same order as the ranking $r$.

\begin{table*}[hbt!]
\centering
\caption{Hyperparameters of fine-tuning our models with elastic dimensions.}
\resizebox{0.8\linewidth}{!}{
\begin{tabular}{c|c|c|c|c|c}
\hline
                                                                                     &                      & LSUN-Bedroom & LSUN-Church & FFHQ   & ImageNet \\ \hline
\multirow{3}{*}{\begin{tabular}[c]{@{}c@{}}Elastic Depth\\ Fine-tuning\end{tabular}} & Batch Size$\times$Num GPU & 32$\times$8       & 32$\times$8      & 32$\times$8 & 32$\times$8     \\ \cline{2-6} 
                                                                                     & learning rate        & 9.6e-5       & 5e-5        & 8.4e-5 & 8e-5     \\ \cline{2-6} 
                                                                                     & Iterations           & 200k         & 50k        & 30k    & 40k      \\ \hline
\multirow{3}{*}{\begin{tabular}[c]{@{}c@{}}Elastic Width\\ Fine-tuning\end{tabular}} & Batch Size$\times$Num GPU & 32$\times$8       & 32$\times$8      & 32$\times$8 & 32$\times$8     \\ \cline{2-6} 
                                                                                     & learning rate        & 9.6e-5       & 5e-5        & 8.4e-5 & 8e-5     \\ \cline{2-6} 
                                                                                     & Iterations           & 130k         & 50k        & 30k    & 40k      \\ \hline
\end{tabular}}
\label{elastic_training_hparams}
\end{table*}

\subsection{Pruning Algorithm}
We present our pruning algorithm to train our Expert Routing Agent to select proper sub-networks of the experts in Alg.\ref{pruning-alg}.

% \begin{algorithm}[t!] 
	
% 	\caption{Our Pruning Algorithm}
% 	\SetAlgoLined
% 	\SetNoFillComment
	
% 	\textbf{Input:}{~Training dataset $\mathcal{D}=\{(x_m, c_m)\}_{m=1}^D$ of images $x_i$ and possible conditional inputs $c_i$; ERA model $h_\text{ERA}(z; \beta)$; Elastically fine-tuned experts $E_i$; pruning iterations $G$; Total MACs budget $T_d$} \\
% 	\textbf{Output:}{~Trained Expert Routing Agent.}

% 	\For{$e:=1$ to $G$}{
		
% 		\State 1. Sample a mini-batch $(\mathbf{x}, \mathbf{c})$ from $\mathcal{D}$.
		
% 		%%%%%%%%%%%%%%%%%%%%%
		
% 		\State 2.~Calculate architecture vectors $(u^{(i)}, v^{(i)})$ using the ERA model $h_\text{ERA}(z; \beta)$ and Eqs.~\ref{v_calculation},~\ref{u_calculation}.

% 		%%%%%%%%%%%%%%%%%%%%%
		
% 		\State 3.~Compute soft pruning vectors $(\mathbf{u}^{(i)}, \mathbf{v}^{(i)})$ using Eqs~(7, 9) in the paper.
		
% 		%%%%%%%%%%%%%%%%%%%%%
		
% 		\State 4.~Apply the soft pruning vectors $(\mathbf{u}^{(i)}, \mathbf{v}^{(i)})$ to the experts using Eqs.~(8, 10, 11) in the paper.
		
% 		%%%%%%%%%%%%%%%%%%%%%
		
% 		\State 5.~Calculate the interval denoising objectives $\mathcal{L}_{\text{DDPM}, \mathcal{I}_i}(E_i(u^{(i)}, v^{(i)}))$ for the experts using the samples $(\mathbf{x}, \mathbf{c})$ in the mini-batch.

% 		%%%%%%%%%%%%%%%%%%%%%
		
% 		\State 6.~Compute the MACs regularization term $\mathcal{R}(\widehat{T}(u, v), T_d)$.

%             \State 7.~Compute the training objective $\mathcal{J}(T_d)$, backpropagate the gradients \emph{w.r.t} the ERA parameters $\beta$ and update them.
		
% 		}
% 	%%%%%%%%%%%%%%%%%%%%%

% 	\textbf{Return:} Trained ERA model.
% 	\label{pruning-alg}
% \end{algorithm}

\begin{algorithm}[t!] 
	\caption{Our Pruning Algorithm}
	\SetAlgoLined
	\SetNoFillComment
	
	\textbf{Input:}{~Training dataset $\mathcal{D}=\{(x_m, c_m)\}_{m=1}^D$ of images $x_i$ and possible conditional inputs $c_i$; ERA model $h_\text{ERA}(z; \beta)$; Elastically fine-tuned experts $E_i$; pruning iterations $G$; Total MACs budget $T_d$} \\
	\textbf{Output:}{~Trained Expert Routing Agent.}
	
	\For{$e:=1$ to $G$}{
		
		1. Sample a mini-batch $(\mathbf{x}, \mathbf{c})$ from $\mathcal{D}$.
		
		2. Calculate architecture vectors $(u^{(i)}, v^{(i)})$ using the ERA model $h_\text{ERA}(z; \beta)$ and Eqs.~\ref{v_calculation},~\ref{u_calculation}.
  
		3. Compute soft pruning vectors $(\mathbf{u}^{(i)}, \mathbf{v}^{(i)})$ using Eqs~\ref{width-gumbell}, \ref{depth-gumbell}.
		
		4. Apply the soft pruning vectors $(\mathbf{u}^{(i)}, \mathbf{v}^{(i)})$ to the experts using Eqs.~\ref{width_pruning_eq}, \ref{depth_pruning_enc}, \ref{depth_pruning_dec}.
		
		5. Calculate the interval denoising objectives $\mathcal{L}_{\text{DDPM}, \mathcal{I}_i}(E_i(u^{(i)}, v^{(i)}))$ for the experts using the samples $(\mathbf{x}, \mathbf{c})$ in the mini-batch.
		
		6. Compute the MACs regularization term $\mathcal{R}(\widehat{T}(u, v), T_d)$.

        7. Compute the training objective $\mathcal{J}(T_d)$, backpropagate the gradients \emph{w.r.t} the ERA parameters $\beta$ and update them.
	}
	\textbf{Return:} Trained ERA model.
	\label{pruning-alg}
\end{algorithm}

\section{Experiments}
We provide more details about our experimental setup as well as experimental results in this section. 

\subsection{Setup}
We implement our method upon the LDM codebase\footnote{\href{https://github.com/CompVis/latent-diffusion}{https://github.com/CompVis/latent-diffusion}} and mainly follow the hyperparameter settings of the LDM~\cite{rombach2022LDM}.~We refer to Tables (12, 13) of LDM \footnote{\href{https://arxiv.org/pdf/2112.10752.pdf}{https://arxiv.org/pdf/2112.10752.pdf}} for hyperparameters of the architecture of the pretrained models that we use in our experiments.~We use the DDIM sampler~\cite{song2021DDIM} for sampling from our pruned models.~We set the number of sampling steps to 100 for the LSUN-Bedroom, 200 for the FFHQ, and 250 for the ImageNet experiments.~We conduct all of our experiments on a server with 8 NVIDIA A100 GPUs.~We calculate the inference throughput value for each model by sampling a batch of 64 samples from it 100 times and averaging the throughput values. We provide more details of our experimental setup for each stage of our method in the following. 

\subsubsection{Fine-tuning with Elastic Dimensions}  Tab~\ref{elastic_training_hparams} summarizes the hyperparameters that we use to fine-tune our experts with elastic depth and width on their intervals.~We adopt the learning rate values from the settings used to train the pre-trained checkpoints in the LDM~\cite{rombach2022LDM} paper.

\subsubsection{Pruning and Fine-tuning}
We provide the hyperparameters for the pruning and fine-tuning stages of our method in Tab.~\ref{pr_ft_hparams}.

For the pruning stage of our method on all datasets, we use the AdamW optimizer~\cite{loshchilov2018AdamW} with a learning rate of $0.001$, weight decay of $0.01$, and beta parameters $(\beta_1, \beta_2) = (0.9, 0.99  9)$ to train the ERA model's parameters.~We also set the temperature parameter $\tau$ for the Gumbell-Sigmoid~\cite{jang2017GumbelSoftmax} estimations to $0.4$ for all experiments.

\subsubsection{Ablation Experiments}
We prune all the baselines in the ablation experiments for 60k iterations.~Then, we match their fine-tuning iterations with the summation of the iterations of our elastic depth fine-tuning, elastic width fine-tuning, and fine-tuning the mixture of efficient experts for the 50\% MACs budget (550k iterations) for a fair comparison.

\begin{table*}[]
\centering

\caption{Hyperparameters for the pruning and fine-tuning stages of our method for different MACs pruning ratios (30\%, 50\%, and 70\%).}
\resizebox{0.7\linewidth}{!}{
\begin{tabular}{c|c|ccc|ccc}
\hline
                              &                    & \multicolumn{3}{c|}{Pruning}                                 & \multicolumn{3}{c}{Fine-tuning}                                    \\ \hline
Dataset                       & Parameters         & \multicolumn{1}{c|}{30\%}  & \multicolumn{1}{c|}{50\%}  & 70\%  & \multicolumn{1}{c|}{30\%}    & \multicolumn{1}{c|}{50\%}    & 70\%    \\ \hline
\multirow{3}{*}{LSUN-Bedroom} & Batch Size$\times$Num GPU & \multicolumn{1}{c|}{12$\times$8} & \multicolumn{1}{c|}{12$\times$8} & 12$\times$8 & \multicolumn{1}{c|}{32$\times$8}   & \multicolumn{1}{c|}{32$\times$8}   & 24$\times$8   \\ \cline{2-8} 
                              & learning rate      & \multicolumn{1}{c|}{-}    & \multicolumn{1}{c|}{-}    & -    & \multicolumn{1}{c|}{9.6e-5} & \multicolumn{1}{c|}{9.6e-5} & 9.6e-5 \\ \cline{2-8} 
                              & Iterations         & \multicolumn{1}{c|}{70k}  & \multicolumn{1}{c|}{60k}  & 50k  & \multicolumn{1}{c|}{270k}   & \multicolumn{1}{c|}{220k}   & 195k   \\ \hline
\multirow{3}{*}{LSUN-Church}  & Batch Size$\times$Num GPU & \multicolumn{1}{c|}{12$\times$8} & \multicolumn{1}{c|}{12$\times$8} & 12$\times$8 & \multicolumn{1}{c|}{32$\times$8}   & \multicolumn{1}{c|}{32$\times$8}   & 24$\times$8   \\ \cline{2-8} 
                              & learning rate      & \multicolumn{1}{c|}{-}    & \multicolumn{1}{c|}{-}    & -    & \multicolumn{1}{c|}{5e-5}   & \multicolumn{1}{c|}{5e-5}   & 5e-5   \\ \cline{2-8} 
                              & Iterations         & \multicolumn{1}{c|}{70k}  & \multicolumn{1}{c|}{60k}  & 50k  & \multicolumn{1}{c|}{165k}   & \multicolumn{1}{c|}{180k}   & 90k    \\ \hline
\multirow{3}{*}{FFHQ}         & Batch Size$\times$Num GPU & \multicolumn{1}{c|}{12$\times$8} & \multicolumn{1}{c|}{12$\times$8} & 12$\times$8 & \multicolumn{1}{c|}{32$\times$8}   & \multicolumn{1}{c|}{32$\times$8}   & 24$\times$8   \\ \cline{2-8} 
                              & learning rate      & \multicolumn{1}{c|}{-}    & \multicolumn{1}{c|}{-}    & -    & \multicolumn{1}{c|}{8.4e-5} & \multicolumn{1}{c|}{8.4e-5} & 8.4e-5 \\ \cline{2-8} 
                              & Iterations         & \multicolumn{1}{c|}{40k}  & \multicolumn{1}{c|}{30k}  & 20k  & \multicolumn{1}{c|}{90k}    & \multicolumn{1}{c|}{85k}    & 100k   \\ \hline
\multirow{3}{*}{ImageNet}     & Batch Size$\times$Num GPU & \multicolumn{1}{c|}{12$\times$8} & \multicolumn{1}{c|}{12$\times$8} & 12$\times$8 & \multicolumn{1}{c|}{32$\times$8}   & \multicolumn{1}{c|}{32$\times$8}   & 24$\times$8   \\ \cline{2-8} 
                              & learning rate      & \multicolumn{1}{c|}{-}    & \multicolumn{1}{c|}{-}    & -    & \multicolumn{1}{c|}{8e-5}   & \multicolumn{1}{c|}{8e-5}   & 8e-5   \\ \cline{2-8} 
                              & Iterations         & \multicolumn{1}{c|}{50k}  & \multicolumn{1}{c|}{40k}  & 30k  & \multicolumn{1}{c|}{205k}   & \multicolumn{1}{c|}{130k}   & 135k   \\ \hline
\end{tabular}
}
\label{pr_ft_hparams}
\end{table*}

\subsection{Comparison of Training Iterations}
We provide a comparison of the number of training iterations for different methods to obtain a pruned model on LSUN-Bedroom and LSUN-Church in Tabs.~\ref{lsun_bed_iters},~\ref{lsun_church_iters} respectively.~For example, our method's total number of iterations is the summation of iterations for pre-training, elastic depth fine-tuning, elastic width fine-tuning, pruning, and fine-tuning the mixture of experts.

Although not directly comparable to SP~\cite{fang2023StructuralPruningforDMs} as it prunes a pixel-space DPM, our method can obtain a pruned model with significantly better quality with less iterations than SP.~This shows that LDMs have much fewer redundancies than pixel-space DPMs.~Thus, they can converge faster and pruning them is more challenging.

On LSUN-Church,~On the one hand, DiffPruning~(70\%) converges with only 0.74M iterations while pixel-space pruned model by SP~\cite{fang2023StructuralPruningforDMs} requires 4.9M iterations to converge with a 4.51 worse FID score.~On the other hand, the comparison results with OMS-DPM~\cite{liu2023OMS-DPM} clearly demonstrate the value of our elastic fine-tuning.~DiffPruning 50\% and 30\% require more than 7$\times$ less training iterations than OMS-DPM to obtain a performant mixture of efficient experts.~The reason is that our elastic fine-tuning scheme provides an implicit model zoo within the experts for each interval without requiring to train multiple models from scratch to obtain a model zoo as done in OMS-DPM~\cite{liu2023OMS-DPM}.

\begin{table}[]
\centering
\caption{Comparison of the number of training iterations for different methods on LSUN-Bedroom.~The ``Method's Iterations" column denotes the number of all the training iterations that the pruning method performs to obtain its final efficient model.}
\resizebox{\linewidth}{!}{
\begin{tabular}{ccccccc}
\hline
\multicolumn{7}{c}{LSUN-Bedroom ($256\times256$)}                                                                                                                                                                                                                                                                                                                                                                                       \\ \hline
\multicolumn{1}{c|}{}             & \multicolumn{4}{c|}{Complexity}                                                                                                                                                                                                                                                              & \multicolumn{2}{c}{Performance}                                                               \\ \hline
\multicolumn{1}{c|}{Model}        & \multicolumn{1}{c|}{\begin{tabular}[c]{@{}c@{}}Pre-training\\ Iterations\end{tabular}} & \multicolumn{1}{c|}{\begin{tabular}[c]{@{}c@{}}Method's\\ Iterations\end{tabular}} & \multicolumn{1}{c|}{\begin{tabular}[c]{@{}c@{}}Total\\ Iterations\end{tabular}} & \multicolumn{1}{c|}{MACs}    & \multicolumn{1}{c|}{\begin{tabular}[c]{@{}c@{}}Throughput $(\uparrow)$\\ (Sample/Sec)\end{tabular}} & FID $(\downarrow)$  \\ \hline
\multicolumn{1}{c|}{DDPM~\cite{ho2020ddpm}}         & \multicolumn{1}{c|}{2.4M}                                                              & \multicolumn{1}{c|}{-}                                                             & \multicolumn{1}{c|}{2.4M}                                                       & \multicolumn{1}{c|}{248.7G}  & \multicolumn{1}{c|}{0.74}                                                              & 6.62 \\
\multicolumn{1}{c|}{SP~\cite{fang2023StructuralPruningforDMs}}           & \multicolumn{1}{c|}{2.4M}                                                              & \multicolumn{1}{c|}{0.2M}                                                          & \multicolumn{1}{c|}{2.6M}                                                       & \multicolumn{1}{c|}{138.8G}  & \multicolumn{1}{c|}{-}                                                                 & 18.6 \\ \hline
\multicolumn{1}{c|}{LDM~\cite{rombach2022LDM}}          & \multicolumn{1}{c|}{1.9M}                                                              & \multicolumn{1}{c|}{-}                                                             & \multicolumn{1}{c|}{1.9M}                                                       & \multicolumn{1}{c|}{101.32G} & \multicolumn{1}{c|}{2.01}                                                              & 4.39 \\
\multicolumn{1}{c|}{DiffPruning (70\%)} & \multicolumn{1}{c|}{1.9M}                                                              & \multicolumn{1}{c|}{0.575M}                                                        & \multicolumn{1}{c|}{2.475M}                                                     & \multicolumn{1}{c|}{70.84G}  & \multicolumn{1}{c|}{3.11}                                                              & 5.90 \\
\multicolumn{1}{c|}{DiffPruning (50\%)} & \multicolumn{1}{c|}{1.9M}                                                              & \multicolumn{1}{c|}{0.61M}                                                         & \multicolumn{1}{c|}{2.51M}                                                      & \multicolumn{1}{c|}{50.69G}  & \multicolumn{1}{c|}{3.75}                                                              & 6.73 \\
\multicolumn{1}{c|}{DiffPruning (30\%)} & \multicolumn{1}{c|}{1.9M}                                                              & \multicolumn{1}{c|}{0.67M}                                                         & \multicolumn{1}{c|}{2.57M}                                                      & \multicolumn{1}{c|}{31.11G}  & \multicolumn{1}{c|}{4.73}                                                              & 9.22 \\ \hline
\end{tabular}
 }
\label{lsun_bed_iters}
\end{table}

\begin{table}[]
\centering
\caption{Comparison of the number of training iterations for different methods on LSUN-Church.~The ``Method's Iterations" column denotes the number of all the training iterations that the pruning method performs to obtain its final efficient model.}
\resizebox{\linewidth}{!}{
\begin{tabular}{ccccccc}
\hline
\multicolumn{7}{c}{LSUN-Church ($256\times256$)}                                                                                                                                                                                                                                                                                                                                                                                        \\ \hline
\multicolumn{1}{c|}{}             & \multicolumn{4}{c|}{Complexity}                                                                                                                                                                                                                                                             & \multicolumn{2}{c}{Performance}                                                                \\ \hline
\multicolumn{1}{c|}{Model}        & \multicolumn{1}{c|}{\begin{tabular}[c]{@{}c@{}}Pre-training\\ Iterations\end{tabular}} & \multicolumn{1}{c|}{\begin{tabular}[c]{@{}c@{}}Method's\\ Iterations\end{tabular}} & \multicolumn{1}{c|}{\begin{tabular}[c]{@{}c@{}}Total\\ Iterations\end{tabular}} & \multicolumn{1}{c|}{MACs}   & \multicolumn{1}{c|}{\begin{tabular}[c]{@{}c@{}}Throughput~$(\uparrow)$\\ (Sample/Sec)\end{tabular}} & FID~$(\downarrow)$   \\ \hline
\multicolumn{1}{c|}{LDM~\cite{rombach2022LDM}}          & \multicolumn{1}{c|}{0.5M}                                                              & \multicolumn{1}{c|}{-}                                                             & \multicolumn{1}{c|}{0.5M}                                                       & \multicolumn{1}{c|}{20.96}  & \multicolumn{1}{c|}{5.19}                                                              & 5.21  \\ \hline
\multicolumn{1}{c|}{DDPM~\cite{ho2020ddpm,song2021DDIM}}         & \multicolumn{1}{c|}{4.4M}                                                              & \multicolumn{1}{c|}{-}                                                             & \multicolumn{1}{c|}{4.4M}                                                       & \multicolumn{1}{c|}{248.7G} & \multicolumn{1}{c|}{0.74}                                                              & 10.58 \\
\multicolumn{1}{c|}{SP~\cite{fang2023StructuralPruningforDMs}}           & \multicolumn{1}{c|}{4.4M}                                                              & \multicolumn{1}{c|}{0.5M}                                                          & \multicolumn{1}{c|}{4.9M}                                                       & \multicolumn{1}{c|}{138.8G} & \multicolumn{1}{c|}{-}                                                                 & 13.9  \\
\multicolumn{1}{c|}{DiffPruning (70\%)} & \multicolumn{1}{c|}{0.5M}                                                              & \multicolumn{1}{c|}{0.24M}                                                         & \multicolumn{1}{c|}{0.74M}                                                      & \multicolumn{1}{c|}{14.64G} & \multicolumn{1}{c|}{5.73}                                                              & 9.39  \\ \hline
\multicolumn{1}{c|}{OMS-DPM~\cite{liu2023OMS-DPM}}      & \multicolumn{1}{c|}{0}                                                                 & \multicolumn{1}{c|}{\textgreater 6M}                                               & \multicolumn{1}{c|}{\textgreater 6M}                                            & \multicolumn{1}{c|}{-}      & \multicolumn{1}{c|}{2.56}                                                              & 11.10 \\
\multicolumn{1}{c|}{DiffPruning (50\%)} & \multicolumn{1}{c|}{0.5M}                                                              & \multicolumn{1}{c|}{0.34M}                                                         & \multicolumn{1}{c|}{0.84M}                                                      & \multicolumn{1}{c|}{10.48G} & \multicolumn{1}{c|}{6.28}                                                              & 10.89 \\ \hline
\multicolumn{1}{c|}{OMS-DPM~\cite{liu2023OMS-DPM}}      & \multicolumn{1}{c|}{0}                                                                 & \multicolumn{1}{c|}{\textgreater 6M}                                               & \multicolumn{1}{c|}{\textgreater 6M}                                            & \multicolumn{1}{c|}{-}      & \multicolumn{1}{c|}{6.4}                                                               & 13.7  \\
\multicolumn{1}{c|}{DiffPruning (30\%)} & \multicolumn{1}{c|}{0.5M}                                                              & \multicolumn{1}{c|}{0.335M}                                                        & \multicolumn{1}{c|}{0.835M}                                                     & \multicolumn{1}{c|}{6.35G}  & \multicolumn{1}{c|}{6.87}                                                              & 11.39 \\ \hline
\end{tabular}
}
\label{lsun_church_iters}
\end{table}

\subsection{Errors in MACs Calculation}
We mentioned in the caption of Tab.~(1) as well as Sec.~(4.1) of the paper that the MACs values reported by SP~\cite{fang2023StructuralPruningforDMs} for the LDM~\cite{rombach2022LDM} models for the ImageNet experiments are inaccurate.~We describe the reason in the following.~SP\cite{fang2023StructuralPruningforDMs} adopts the `flops-counter.pytorch'~package\footnote{\href{https://github.com/sovrasov/flops-counter.pytorch}{https://github.com/sovrasov/flops-counter.pytorch}} to measure models' MACs.~This package defines a hook for each of the standard PyTorch~\cite{paszke2019pytorch} layers like $\texttt{nn.Conv2d}$ and keeps a \href{https://github.com/VainF/Diff-Pruning/blob/da894a301a5c0f7aaaec727c32e098001627dd60/exp_code/torch_pruning/utils/op_counter.py#L248}{mapping dictionary} between standard PyTorch layers and their hooks.~The package calculates the MACs of the model by performing the forward pass of the model with a random input and counting the layers' MACs using the defined hooks.~Now,~SP~\cite{fang2023StructuralPruningforDMs} implements the \href{https://github.com/VainF/Diff-Pruning/blob/da894a301a5c0f7aaaec727c32e098001627dd60/diffusers/models/attention_processor.py#L36}{Attention} layer in the U-Net architecture of LDM manually, and the defined Attention module is not an element of the \href{https://github.com/VainF/Diff-Pruning/blob/da894a301a5c0f7aaaec727c32e098001627dd60/exp_code/torch_pruning/utils/op_counter.py#L248}{mapping dictionary} for the MACs calculation hooks.~Thus, the package does not count the number of MACs for the \href{https://github.com/VainF/Diff-Pruning/blob/da894a301a5c0f7aaaec727c32e098001627dd60/diffusers/models/attention_processor.py#L729}{scaled dot product} attention operation as it is not a native PyTorch layer. For instance, SP reports that (Tab. (3) in SP~\cite{fang2023StructuralPruningforDMs}) the LDM model for ImageNet has 99.8G MACs. However, we manually implemented counting the MACs for the attention layers and found that the model actually has 108.78G MACs.

We found a similar problem in the numbers reported by SD~\cite{yang2023DPMsMadeSlim} .~For instance, SD reports that the LDM for LSUN-Church has 18.7G MACs.~We could reproduce the same number when directly using the `flops-counter.pytorch' package.~Yet, we found that the model actually has 20.96G MACs after adding the attention layers' MACs.

\section{Related Work}
\textbf{Mixture of Experts (MoE) Diffusion Models:} MoE methods cluster denoising time-steps of DPMs into intervals and train a separate~\textit{expert} model for each.~eDiff-I~\cite{balaji2022ediffiMOE} supports developing MoE for DPMs by showing that different denoising time-steps have distinct roles.~Yet, how to cluster time-steps is non-trivial.~eDiff-I employs a complex tree-based-branching scheme to divide the denoising path into two intervals sequentially and initializes a child model by its parent.~ERNIE-ViLG~\cite{feng2023ernieMOE} and MEME~\cite{lee2023MEME} uniformly cluster the denoising time-steps.~Yet, these heuristic schemes do not necessarily transfer to other tasks.~Different from these methods,~we propose to cluster the denoising time-steps by measuring the alignment between their training objectives.~We emphasize that although a recent work~\cite{go2023NegativeTransfer} has explored the time-steps' alignment scores \textit{in the course of training}, our paper is the first one to leverage them to cluster the time-steps for MoE DPMs.

\noindent\textbf{Efficient DPMs.}~The majority of ideas for improving DPMs' efficiency reduce their denoising steps by faster samplers~\cite{lu2023dpmsolver++,zhang2023ExponentialIntegrator,xu2023RestartSampling}, distillation~\cite{salimans2022ProgressiveDistillation,meng2023distillation,habibian2023clockwork}, better noise schedules~\cite{zheng2023TruncatedDPMs,song2021DDIM,nichol2021improvedDDPM,zhang2023gddim,kingma2021variationalDMs,bao2022AnalyticDPM}, learning denoising timesteps to use~\cite{watson2022LearningFastSamplers,watson2022learning}, and caching~\cite{ma2023DeepCache}.~We explore an orthogonal direction, compressing the architecture of DPMs.~LSGM~\cite{vahdat2021score} and LDM~\cite{rombach2022LDM} perform the diffusion process in a lower dimensional latent space of an encoder-decoder pair~\cite{kingma2021variationalDMs,esser2021vqgan}, thereby enjoying significantly faster sampling than pixel-space DPMs.

A few ideas have recently addressed compressing DPMs' architectures having two main categories.~\textbf{Single-model} methods develop a single efficient model for all denoising timesteps.~Structural Pruning (SP)~\cite{fang2023StructuralPruningforDMs} approximates weights' importance using the Taylor expansion and removes structures with low scores.~Yet, SP's performance has been mainly verified on pixel space DPMs, and its pruned models on datasets like LSUN-Church~\cite{yu15lsun} still have more than $6\times$ MACs than the full-size LDM~\cite{rombach2022LDM}.~MobileDiffusion~\cite{zhao2023MobileDiffusion} introduces heuristics to enhance DPMs' efficiency and develops two efficient architectures.~Nevertheless, it is highly non-trivial how to generalize the heuristics for different compute budgets.~Spectral Diffusion (SD)~\cite{yang2023DPMsMadeSlim} introduces a wavelet gating operator and performs frequency domain distillation from a teacher model into a small LDM.~However, the main weakness of single-model methods is that they use the same model for all denoising steps, which is shown to be sub-optimal~\cite{balaji2022ediffiMOE,go2023NegativeTransfer}.~\textbf{Mixture of expert} methods employ a separate model for different stages of the denoising process.~OMS-DPM~\cite{liu2023OMS-DPM} trains a model zoo with various sizes and searches for a proper model schedule given a desired compute budget.~Yet, gathering a model zoo is very costly on large-scale datasets, making OMS-DPM impractical for them.~T-Stich~\cite{pan2024TStich} stitches several models with different sizes, each performing a part of the denoising process. But, similar to OMS-DPM~\cite{liu2023OMS-DPM}, it requires several pretrained models of various sizes, making it costly for practical scenarios.~MEME~\cite{lee2023MEME} and TMDA~\cite{zhang2023TailoredMultiDecoder} cluster denoising timesteps and design a distinct expert for each.~However, they need to manually allocate the compute budget between experts and re-design the experts for a new budget, which makes them cumbersome in practice.~We propose to prune an LDM into a mixture of efficient experts.~We cluster denoising timesteps into intervals using their alignment scores.~Then, we fine-tune the pre-trained model with elastic dimensions on each interval to obtain our experts.~Thus, our method gathers an \textit{implicit} model zoo \emph{within} each expert with much lower training iterations than OMS-DPM.~Finally, we prune all experts simultaneously using our expert routing agent to obtain our mixture of efficient experts.~By doing so, in contrast with MEME~\cite{lee2023MEME} and TMDA~\cite{zhang2023TailoredMultiDecoder}, our method automatically allocates the compute resource (\textit{e.g.,} MACs) between experts.

\noindent\textbf{Model pruning and architecture search.} Our work is also related to model pruning~\cite{li2017PruningFilters,han2015WeightPruning,Molchanov2019TaylorPruning,he2018AutomaticModelCompression,ganjdanesh2022ISP,ganjdanesh2024RLAL,ganjdanesh2024MGGC} and Neural Architecture Search (NAS)~\cite{zoph2017NAS,liu2018DARTS,Cai2020OnceForAll,yao2021JointDetNAS,hou2020DynaBERT,ganjdanesh2023effconv} methods that prune the pretrained models and design novel architectures given a set of computational constraints.~These ideas mainly focus on developing new architectures for image classification tasks, while we aim to design a novel pruning method for latent diffusion models~\cite{rombach2022LDM}.~We refer to recent surveys~\cite{white2023NAS1000Papers,YangSurveyStructuredPruning,cheng2023SurveyDNNPruning} for a detailed reviewing of pruning and NAS methods.

\end{document}